\pdfoutput=1

\documentclass[11pt]{article}

\usepackage[preprint]{acl}

\usepackage{times}
\usepackage{latexsym}
\usepackage{algorithmicx}
\usepackage{algcompatible}

\usepackage[T1]{fontenc}

\usepackage[utf8]{inputenc}

\usepackage{microtype}
\usepackage{enumitem}
\usepackage{bigfoot}
\usepackage{amsmath, amssymb}
\usepackage{subfigure}
\usepackage[linesnumbered,ruled,vlined]{algorithm2e}
\usepackage{multirow}
\usepackage{booktabs}
\usepackage{inconsolata}
\newcommand\blfootnote[1]{%
  \begingroup
  \renewcommand\thefootnote{}\footnote{#1}%
  \addtocounter{footnote}{-1}%
  \endgroup
}
\usepackage{graphicx}

\title{Steering LVLMs via Sparse Autoencoder for Hallucination Mitigation}

\author{
\textbf{Zhenglin Hua\textsuperscript{1,2}$^\dagger$},
\textbf{Jinghan He\textsuperscript{3,4}$^\dagger$},
\textbf{Zijun Yao\textsuperscript{5}},
\textbf{Tianxu Han\textsuperscript{6}}, \\
\textbf{Haiyun Guo\textsuperscript{3,4}$^\ast$},
\textbf{Yuheng Jia\textsuperscript{1,2}$^\ast$},
\textbf{Junfeng Fang\textsuperscript{7}$^\ast$} \\
\small{\textsuperscript{1}School of Computer Science and Engineering, Southeast University} \\
\small{\textsuperscript{2}Key Laboratory of New Generation Artificial Intelligence Technology and Its Interdisciplinary Applications (Southeast University)} \\
\small{\textsuperscript{3}Foundation Model Research Center, Institute of Automation, Chinese Academy of Sciences} \\
\small{\textsuperscript{4}School of Artificial Intelligence, University of Chinese Academy of Sciences} \\
\small{\textsuperscript{5}Department of Computer Science and Technology, Tsinghua University} \\
\small{\textsuperscript{6}Wuhan University of Technology} \small{\textsuperscript{7}National University of Singapore} \\
\small{\texttt{huazhenglin2003@gmail.com}, \texttt{hejinghan2022@ia.ac.cn}}
}

\begin{document}
\maketitle
\blfootnote{$^\dagger$ Equal contribution.}
\blfootnote{$^\ast$ Corresponding author.}
\begin{abstract}
Large vision-language models (LVLMs) have achieved remarkable performance on multimodal tasks. However, they still suffer from hallucinations, generating text inconsistent with visual input, posing significant risks in real-world applications. Existing approaches to address this issue focus on incorporating external knowledge bases, alignment training, or decoding strategies, all of which require substantial computational cost and time. Recent works try to explore more efficient alternatives by adjusting LVLMs' internal representations. Although promising, these methods may cause hallucinations to be insufficiently suppressed or lead to excessive interventions that negatively affect normal semantics. In this work, we leverage sparse autoencoders (SAEs) to identify semantic directions closely associated with faithfulness or hallucination, extracting more precise and disentangled hallucination-related representations. Our analysis demonstrates that interventions along the identified faithful direction can mitigate hallucinations, while those along the hallucinatory direction can exacerbate them. Building on these insights, we propose \textbf{S}teering LVLMs via \textbf{S}AE \textbf{L}atent Directions (SSL), a plug-and-play method based on SAE-derived latent directions to mitigate hallucinations in LVLMs. Extensive experiments demonstrate that SSL significantly outperforms existing decoding approaches in mitigating hallucinations, while maintaining transferability across different model architectures with negligible additional time overhead. The code is available at \href{https://github.com/huazhenglin2003/SSL}{https://github.com/huazhenglin2003/SSL}.
\end{abstract}

\begin{figure}[t]
  \includegraphics[width=\columnwidth]{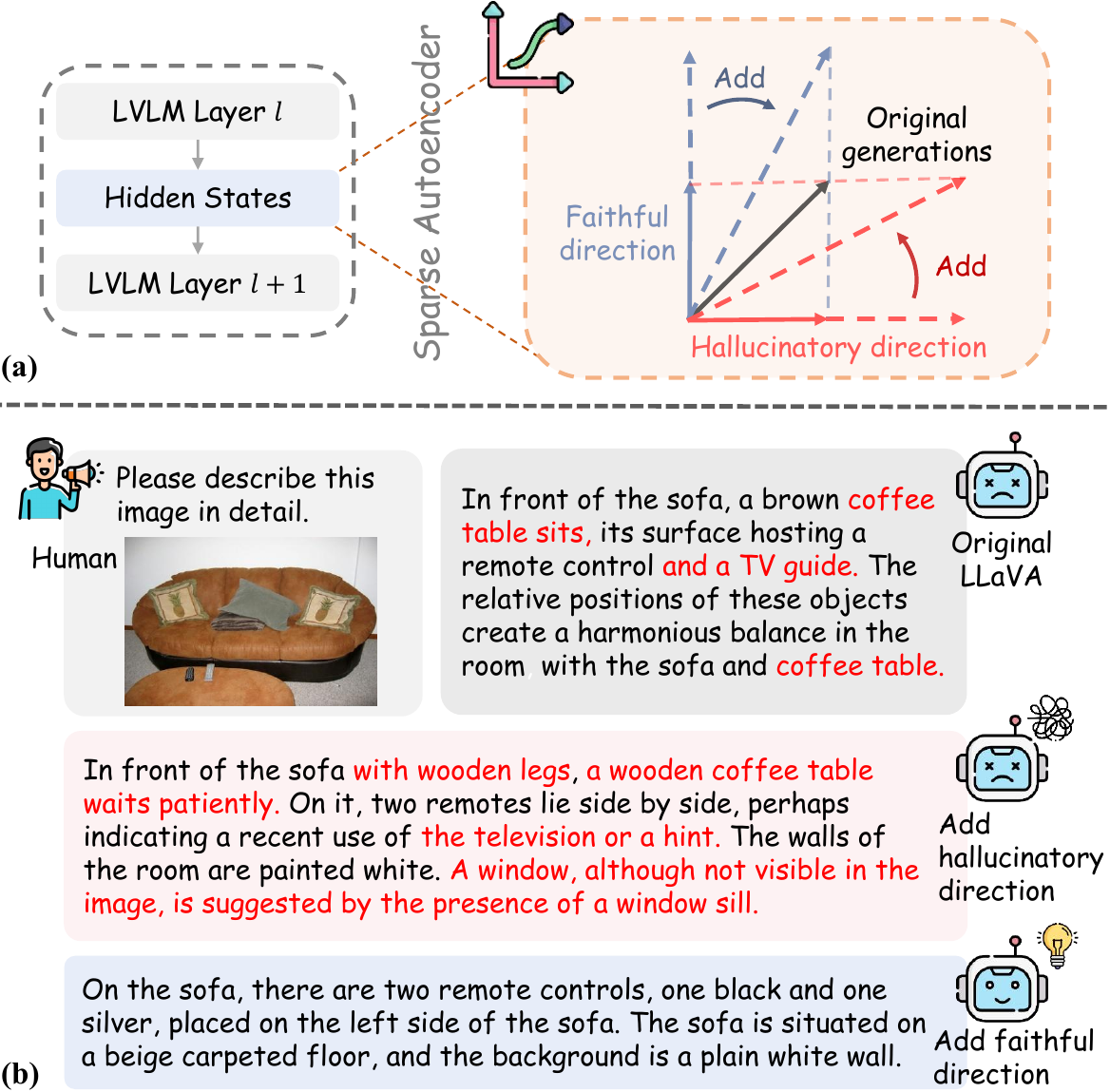}
  \caption{The figure shows, from top to bottom, the original response from the LLaVA-Next-8b, the response after intervention along the hallucinatory direction, and the response after intervention along the faithful direction. These results demonstrate that targeted interventions along faithful directions suppress hallucinatory generations, while perturbations along hallucinatory directions tend to elicit more factually incorrect content.}
  \label{introduction}
\end{figure}

\section{Introduction}

LVLMs \cite{liu2023visualinstructiontuning,dai2023instructblipgeneralpurposevisionlanguagemodels,10655294} have demonstrated impressive capabilities in jointly processing visual and textual modalities, achieving strong performance on tasks such as VQA \cite{vqa} and image captioning \cite{li2022blip}. However, LVLMs still suffer from hallucination \cite{hallucination}, where the generated text does not align with the visual content. This limitation poses significant challenges to their deployment in critical applications, including medical diagnosis\cite{gu2024medvhsystematicevaluationhallucination} and autonomous driving\cite{autonomous}, where factual consistency and reliability are essential.

To mitigate hallucination in LVLMs, researchers explore various strategies, including refining decoding algorithms \cite{VCD,OPERA,code}, incorporating external knowledge bases \cite{qu2024alleviatinghallucinationlargevisionlanguage}, and leveraging additional annotated data for model fine-tuning \cite{park2024mitigatingdialoguehallucinationlarge}. While promising, these approaches often incur substantial computational cost and time. Recent works \cite{VTI,jiang2025interpretingeditingvisionlanguagerepresentations,li2025hiddenlifetokensreducing} try to explore more efficient alternatives by adjusting LVLMs’ internal representations. However, these methods may cause hallucinations to be insufficiently suppressed or lead to excessive interventions that negatively affect normal semantics. Therefore, extracting fine-grained and reliable representations related to hallucinations remains a key challenge in advancing the reliability of LVLMs.

Recently, SAEs have shown success in extracting fine-grained semantic representations—specifically capturing whether the model knows certain entities—of abstract concepts in the field of large language models (LLMs) \cite{doiknow}. SAEs builds on the Linear Representation Hypothesis \cite{park2023the}, which posits that internal model representations can be expressed as sparse combinations of interpretable semantic directions \cite{hollinsworth-etal-2024-language,li2023inferencetime}. Building on this foundation, \citet{sae} utilize SAEs to interpret LVLM representations and analyze open-set semantics. However, their work does not address hallucination modeling or intervention. Inspired by these works, we extend the application of SAE-based analysis from LLMs to LVLMs, with a particular focus on mitigating hallucinations. Specifically, we leverage the SAE provided by \citet{sae} to identify latent directions correlated with hallucinatory semantics as well as those aligned with faithful content, and steer the internal semantic representations of LVLMs, aiming to understand and mitigate hallucinations more precisely and directly. As illustrated in Figure~\ref{introduction}, targeted interventions along faithful directions suppress hallucinatory generations, while perturbations along hallucinatory directions tend to elicit more factually incorrect content.

Building on this insight, we propose \textbf{S}teering LVLMs via \textbf{S}AE \textbf{L}atent Directions (SSL), a plug-and-play approach based on SAE-derived latent directions to mitigate hallucinations in LVLMs. During the visual feature merging stage, we inject faithful semantic directions to amplify grounded semantic features and improve image–text consistency. In the subsequent language generation stage, we reduce projection onto hallucinatory semantic directions, thereby reducing the risk of generating factually incorrect content. Remarkably, although the SAE was trained on the LLaVA-Next-8b model, the extracted hallucination and factuality directions generalize seamlessly to other architectures (\textit{e.g.}, LLaVA1.5-7b model \cite{llava1.5} and InstructBLIP-7b model \cite{Instructblip}). Experimental evaluation on established LVLM hallucination benchmarks shows that SSL outperforms existing decoding approaches, confirming its effectiveness and efficiency in hallucination reduction.

Our main contributions are as follows:

\begin{enumerate}[label=\textbullet]
    \item We leverage SAEs to identify semantic directions that are highly correlated with hallucinatory and faithful content in the representation space of LVLMs. 
    \item We propose SSL, a plug-and-play method that injects factuality semantic directions during visual feature fusion to reinforce grounded content and suppresses hallucination directions during language generation to proactively mitigate hallucinatory outputs.
    \item Extensive experiments demonstrate that SSL outperforms existing decoding approaches on widely used hallucination benchmarks with negligible time overhead, exhibiting transferability across different architectures. 
\end{enumerate}

\section{Preliminary}

\paragraph{LVLM generation.}LVLMs take both image and text as input and encode them into a sequence of tokens. During autoregressive generation, the model first concatenates the system tokens $X_s$, prompt tokens $X_t$, and visual tokens $X_v$ in a predefined order to form the initial input. At the first generating step $t=1$, the model predicts the output token based on this initial context. At each subsequent step $t>1$, the previously generated tokens $X_o^{<t}$ are appended to the end of the initial input, resulting in the current sequence $\left[X_s, X_t, X_v, X_o^{<t}\right]$. The model then generates the next token autoregressively according to the conditional probability distribution, continuing until an end-of-sequence token is produced or a maximum sequence length is reached:
\begin{equation}
    y_t = \arg\max p_\theta(y_t\mid X_s, X_t, X_v,X_o^{<t}),
\end{equation}
where $y_t$ is the token generated at time step $t$.

\paragraph{Sparse autoencoders.} SAEs have been proven to be effective for separating overlapping features \cite{bricken2023monosemanticity,doiknow}. In this work, we use the SAE provided by \citet{sae}, which operates on the residual stream $h_l\in\mathbb{R}^d$ from the $l$-th layer of LVLMs. The SAE projects these representations into a higher-dimensional latent space $z(x)\in\mathbb{R}^{d_{\text{SAE}}}$ and applies a ReLU activation:
\begin{equation}
    z(x)=\text{ReLU}(W_{\text{enc}}x+b_{\text{enc}}),
\end{equation}
where $W_{\text{enc}}$ and $b_{\text{enc}}$ denote the encoder's weight matrix and bias, respectively. To enforce sparsity, a top-k operation retains only the $k$ largest activations in $z(x)$, zeroing out the rest to obtain the sparse latent vector $z_k(x)= \text{TopK}(z, k)$. The decoder then reconstructs the original representation via a linear combination of the active components:
\begin{equation}
    \text{SAE}(x)=W_{\text{dec}}^Tz_k(x) + b_{\text{dec}},
\end{equation}
where $W_{\text{dec}}$ and $b_{\text{dec}}$ denote the decoder’s weight matrix and bias. During training, the loss function combines the reconstruction error with an auxiliary loss proposed by \citet{topk-sae}, aiming to encourage the utilization of meaningful features in the latent representation $z_k(x)$ and to prevent feature inactivity, thereby enhancing the overall expressiveness of the sparse encoding. We refer to each component of $z_k(x)$ as a latent activation, and each row vector of $W_\text{dec}$ as a latent direction.
\begin{figure}[t]
\subfigure[Hallucinatory latent KDE distribution]{\includegraphics[width=0.47\columnwidth]{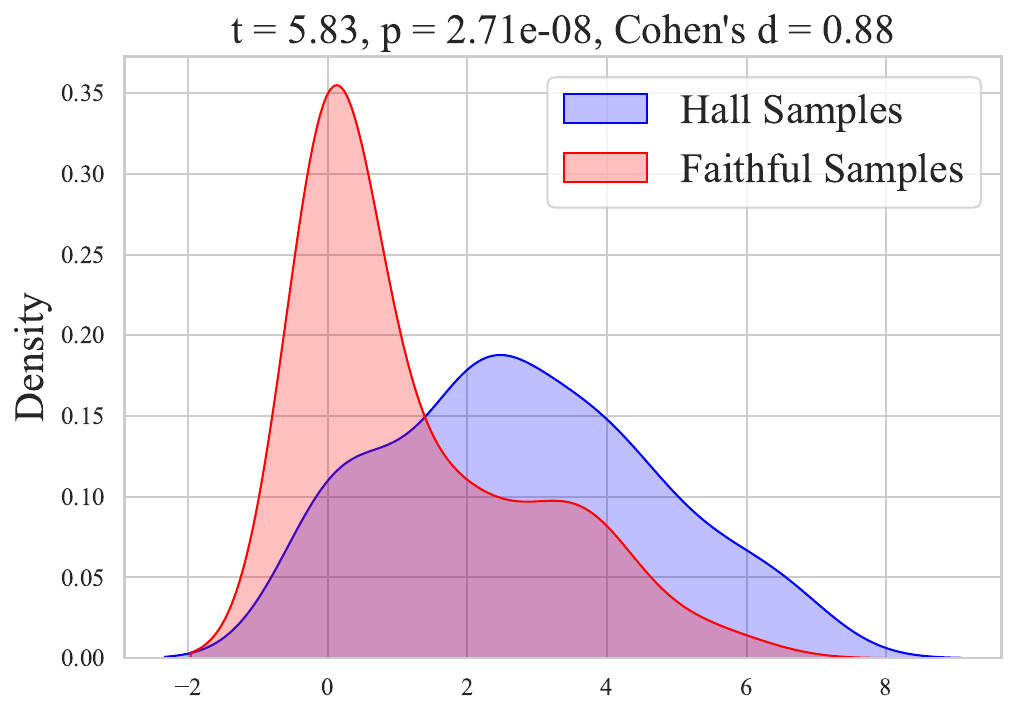}}
\hfill
\subfigure[Faithful latent KDE distribution]{\includegraphics[width=0.47\columnwidth]{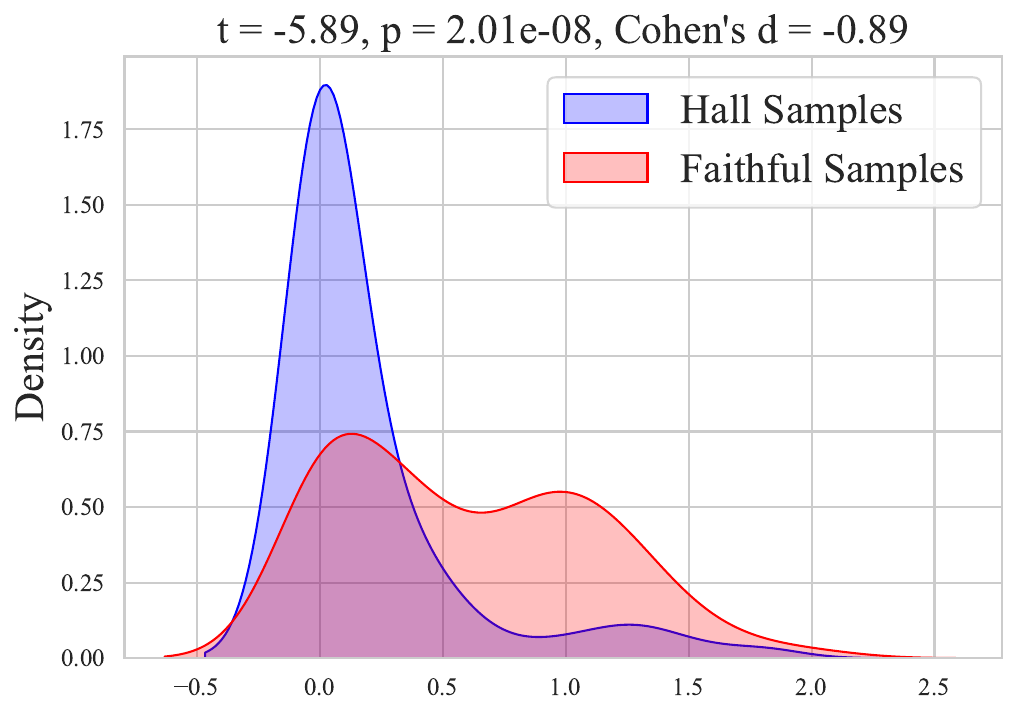}}
  \caption{KDE plots of the selected latent activations for test samples labeled as hallucination and faithfulness. The overlaid plots visualize the distributional differences, with annotated $t$-statistic, $p$-value, and Cohen’s $d$ effect size indicating the statistical separation between the two groups.}
  \label{kde}
\end{figure}

\paragraph{Steering with SAE Latents.}\label{Steer} The SAE reconstructs model representations as a linear combination of latent directions and a bias, effectively approximating the original input. Each latent activation $z_j(x)$ corresponds to a specific decoder direction $d_j=W_{\text{dec}}\left[j,:\right]$, enabling targeted adjustment of the representation through activation steering \cite{DBLP:journals/corr/abs-2308-10248}. This technique allows us to steer the residual stream by modifying the representation as follows:
\begin{equation}
    x_{\text{steer}} \leftarrow x + \alpha d_j,
\label{alpha}
\end{equation}
where $\alpha$ is a tunable parameter that determines the strength of the intervention.

\section{Method}

In this work, we introduce SSL, a plug-and-play method for steering LVLMs. Our method consists of two principal components: semantic directions identification and steering LVLMs via SAE latents.
\subsection{Semantic Directions Identification }
\paragraph{Residual Stream Dataset Construction for Hallucinations and Faithfulness.}To investigate whether there exist directions in LVLMs that are highly correlated with hallucinatory and faithful semantics, we randomly sampled $4,000$ image-text pairs from the MSCOCO dataset \cite{MSCOCO}. Using the LLaVA-Next-8b model \cite{liu2024llavanext} for the image captioning task, we extract the residual stream representations from the $25^\text{th}$ layer when the model generated object tokens classified as either hallucinatory or faithful.

Given that a small proportion of object words are tokenized into multiple subword units, we exclude these cases to simplify the analysis. Furthermore, because each sample exhibited an imbalance between the counts of hallucinatory and faithful object terms, we enforce class balance by sampling an equal number of residual vectors from each category per image–text pair. See Appendix~\ref{appendix A} for a description of the process. Finally, we construct a balanced dataset containing $1,784$ samples and divide it into a training set and a test set in a $9:1$ ratio, for direction mining and direction effectiveness validation, respectively.

\begin{figure*}[t]
\includegraphics[width=\linewidth]{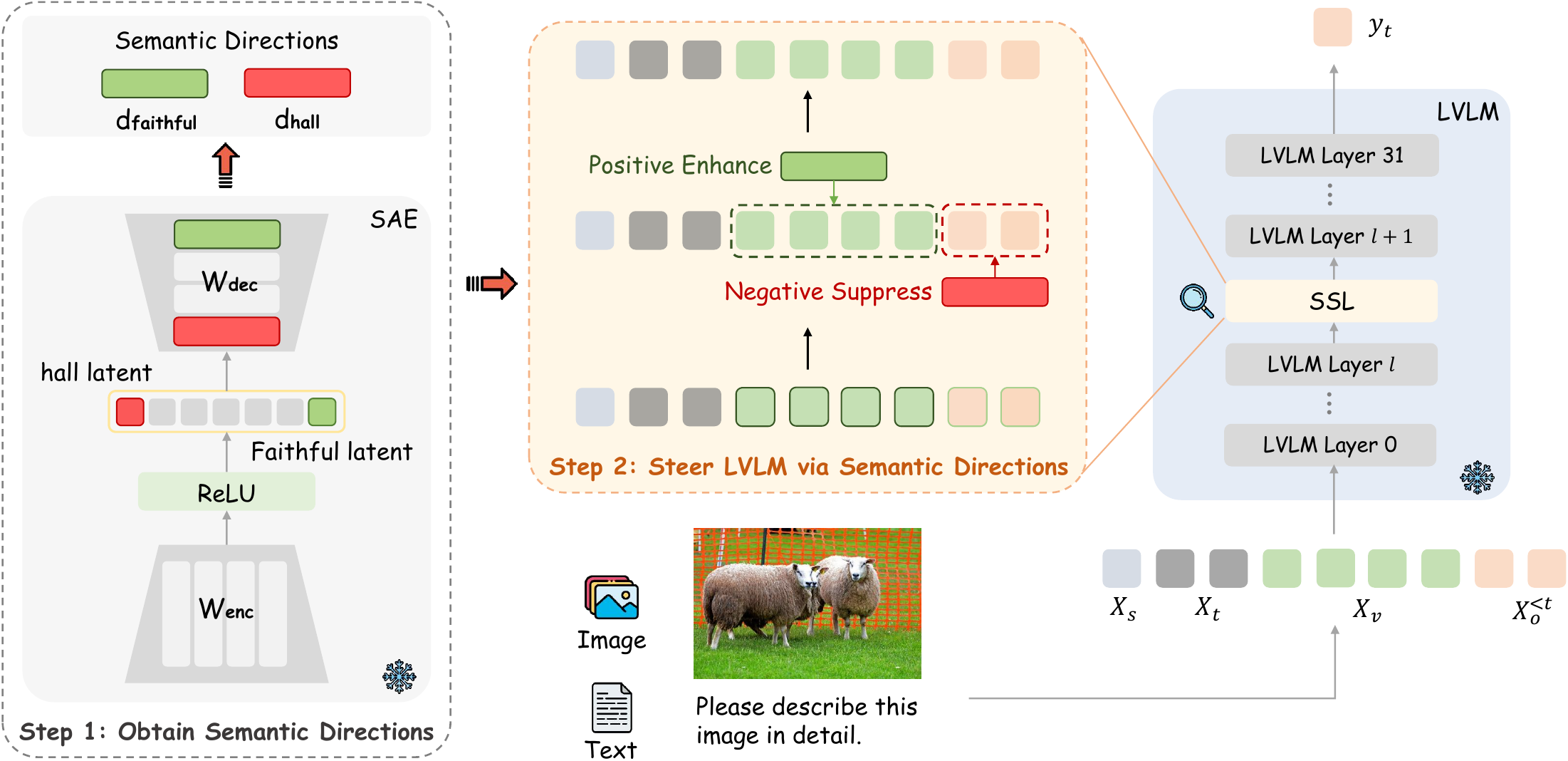}
  \caption{Overview of the proposed SSL approach leveraging SAE to identify semantically aligned directions and mitigate hallucination in LVLMs. We use SAE to identify latent directions within the internal representation space of LVLMs that are associated with hallucinatory and faithful semantics, denoted as $d_{\text{hall}}$ and $d_{\text{faithful}}$, respectively. These semantic directions are then used to modulate the residual stream at the $l$-th layer, steering the generation process toward greater factual consistency.}
  \label{SSL}
\end{figure*}

\paragraph{Semantically Hallucinatory and Faithful Direction Identification via SAE.}Inspired by \citet{meng2022locating,doiknow}, we leverage SAE to identify latent directions aligned with hallucinatory and faithful semantics. Specifically, each residual stream sample from the training set is passed through the SAE, and we record the activation frequency of each latent activation across hallucinatory samples $\mathcal{X_{\text{hall}}}$  and faithful samples $\mathcal{X_{\text{faithful}}}$. For a given latent activation $j$, its activation frequencies on hallucinatory samples $f_j^{\text{hall}}$ and faithful samples $f_j^{\text{faithful}}$ are computed as:
\begin{equation}
\begin{aligned}
f_j^{\text{hall}} &= \frac{1}{N_{\text{hall}}} \sum_{x\in\mathcal{X_{\text{hall}}}} \mathbb{I} \left(z_j(x) > 0 \right), \\
f_j^{\text{faithful}} &= \frac{1}{N_{\text{faithful}}} \sum_{x\in\mathcal{X_{\text{faithful}}}} \mathbb{I} \left(z_j(x) > 0 \right),
\end{aligned}
\end{equation}
where $N_{\text{hall}}$ and $N_{\text{faithful}}$ represent the number of hallucinatory and faithful samples, respectively. To quantify the semantic relevance of each latent activation, we compute the difference in activation frequencies as follows:
\begin{equation}
\begin{aligned}
s_j^{\text{hall}} &= f_j^{\text{hall}} - f_j^{\text{faithful}},\\
s_j^{\text{faithful}} &= f_j^{\text{faithful}} - f_j^{\text{hall}}.
\end{aligned}
\end{equation}
These values reflect the relevance of latent dimension $j$ to hallucinatory and faithful semantics, respectively. Finally, we identify the latent activation with the highest $s_j^{\text{hall}}$ as the hallucination semantic direction (hereafter referred to as the \textit{hall} latent), and the one with the highest $s_j^{\text{faithful}}$ as the faithful semantic direction (hereafter referred to as the \textit{faithful} latent).

\paragraph{Validation of the Effectiveness of Semantic Directions.}We begin by analyzing the distributional differences of the hallucinatory latent and faithful latent activations across both sample types in the test set. These distributions are visualized using kernel density estimation (KDE) plots as shown in Figure~\ref{kde}. We further quantify the separation using independent two-sample $t$-tests and compute Cohen’s $d$ to assess effect sizes. Both latent activations exhibit statistically significant distributional shifts, with substantial effect sizes, confirming their discriminative power.

To probe the semantic alignment of the activations, we further conduct Spearman rank correlation analysis between the activation values of hallucinatory samples and their associated hallucinatory object terms. The hallucinatory latent correlates positively with hallucinatory objects (Spearman’s $\rho = 0.42$, $p = 9.95\times10^{-9}$), whereas the faithful latent correlates negatively ($\rho = -0.44$, $p = 9.85\times10^{-10}$). Given the binary nature of the labels and the symmetry of rank correlation, we infer a positive correlation between the faithful latent and samples.

\begin{figure}[t]
\includegraphics[width=\columnwidth]{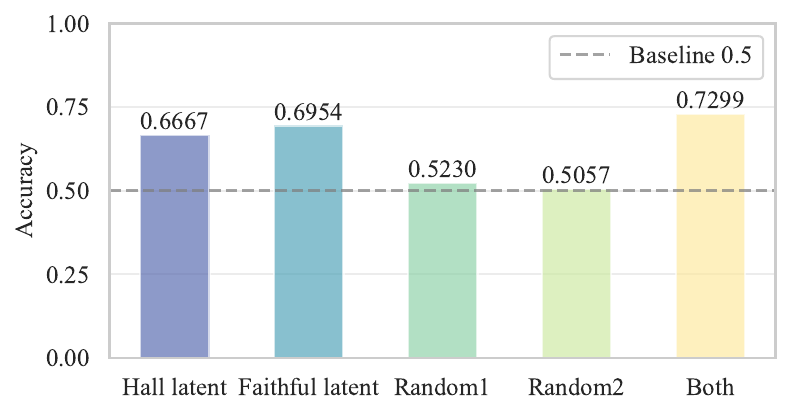}
  \caption{Comparison of classification accuracy using different latent activations. Hall latent and faithful latent correspond to the identified hallucinatory and faithful latent activations, respectively. Random1 denotes a single latent activation randomly selected from the SAE latent space, while Random2 represents a feature combination of two randomly selected latent activations. The dashed line indicates the baseline accuracy of 0.5.}
  \label{logitclassification}
\end{figure}

Finally, to quantitatively evaluate the predictive power of SAE-derived directions in distinguishing hallucinatory from faithful samples, we design a set of classification experiments based on logistic regression. The model takes the latent activation values—either individually or in combination—as input features for a binary hallucination classification task. As illustrated in Figure~\ref{logitclassification}, the results demonstrate that the latent semantic directions extracted by the SAE are discriminative (see Appendix~\ref{appendix B} for more experimental details). Furthermore, combining the \textit{hall} latent and \textit{faithful} latent as input features yields further performance improvements. A more detailed discussion on clarifying the boundary between hallucinatory and faithful semantic directions can be found in Appendix~\ref{appendix C}.

\subsection{Steering LVLMs via SAE Latents}

\paragraph{Steering Strategy.}A core architectural mechanism involves multi-layer residual connections that progressively model input information. At each layer, semantic representations are passed through residual flows, which can be divided according to the input sequence into four main components: system token, prompt tokens, visual tokens, and output tokens. Among these, visual tokens interact with language tokens to guide the model’s understanding of image content. Injecting faithful direction at this position enables the model to increase visual faithfulness. Output tokens represent the model’s autoregressive language generations, influenced by both prompts and visual inputs, thereby reflecting the model’s semantic behavior. Suppressing hallucinatory directions during this stage helps reduce hallucination tendencies and enhances factual consistency in language generation.

Following the method described in Section~\ref{Steer}, we identify two semantic direction vectors: the hallucinatory direction $d_{\text{hall}}$ and the faithful direction $d_{\text{faithful}}$. During the visual feature fusion stage, we incorporate $d_{\text{faithful}}$ to improve the faithfulness of visual understanding. In the subsequent language generation stage, we suppress activations along $d_{\text{hall}}$ to reduce the risk of hallucinatory outputs. Semantic steering at layer $l$ is defined as follows:
\begin{equation}
\begin{aligned}
    X_{l,\text{v}} &\leftarrow X_{l,\text{v}} + \alpha\cdot d_{\text{faithful}},\\
    X_{l,\text{o}}^{<t} &\leftarrow X_{l,\text{o}}^{<t}-\alpha\cdot d_{\text{hall}},
\end{aligned}
\end{equation}
where $\alpha$ is a tunable hyperparameter controlling the strength of semantic steering.

\paragraph{Adaptive Steering Parameters (ASP).}The setting of the steering strength $\alpha$ plays a crucial role in determining the effectiveness of semantic intervention. Traditional steering approaches often rely on a fixed hyperparameter $\alpha$ to linearly combine the steering vector with the residual representations. However, this fixed strategy can result in unstable or suboptimal performance, as the magnitude of residual vectors can vary across model layers and token positions. In such cases, a change that is too small may fail to induce meaningful guidance, while an excessively large change may cause semantic distortion or instability.

To address this limitation, we propose an adaptive feature steering mechanism, which dynamically adjusts the steering strength based on the norm of the residual vector at each token at a given layer. This approach ensures more stable and context-aware intervention across varying model states. Specifically, the adaptive steering strength $\alpha$ is computed as:
\begin{equation}
\alpha = \gamma \cdot \frac{\left \| x_{\text{residual}} \right \| }{\left \| d_{\text{steer}} \right \| +\epsilon},
\end{equation}
where $\gamma$ is a scaling factor, $x_{\text{residual}}$ denotes the residual vector, $d_{\text{steer}}$ is the steering direction, and $\epsilon$ is a small constant to avoid numerical instability. An overview of the proposed SSL method is presented in Figure~\ref{SSL}. The complete procedure of SSL is provided in Algorithm~\ref{alg:ssl}.

\begin{algorithm}[t]
\caption{SSL}
\label{alg:ssl}
\KwIn{Scaling factor $\gamma$; steering layer $l_s$; semantic directions $d_{\text{hall}}$, $d_{\text{faithful}}$; residual stream at layer $l_s$: $[X_{l_s,s}, X_{l_s,t}, X_{l_s,v}, X_{l_s,o}^{<t}]$}
\If{$L = l_s$}{
    \For{token $x$ in residual stream}{
        \uIf{$x \in X_{l_s,v}$}{
            $x \leftarrow x + \gamma \cdot \frac{\|x\|}{\|d_{\text{faithful}}\| + \epsilon} \cdot d_{\text{faithful}}$
        }
        \uElseIf{$x \in X_{l_s,o}^{<t}$}{
            $x \leftarrow x - \gamma \cdot \frac{\|x\|}{\|d_{\text{hall}}\| + \epsilon} \cdot d_{\text{hall}}$
        }
        \Else{
            \tcp{System and prompt tokens remain unchanged}
        }
    }
}
\end{algorithm}

\section{Experiments}

\subsection{LVLMs}

\begin{table*}[t]
\centering
\small
\resizebox{\textwidth}{!}{
\begin{tabular}{@{}c|ccc|ccc|ccc@{}}
\toprule
\multirow{2}{*}{} 
& \multicolumn{3}{c|}{LLaVA-NeXT-8b} 
& \multicolumn{3}{c|}{LLaVA-1.5-7b} 
& \multicolumn{3}{c}{InstructBLIP-7b} \\
& $\text{CHAIR}_S$ $\downarrow$ & $\text{CHAIR}_I$ $\downarrow$ & \textit{Avg.Len} 
& $\text{CHAIR}_S$ $\downarrow$ & $\text{CHAIR}_I$ $\downarrow$ & \textit{Avg.Len} 
& $\text{CHAIR}_S$ $\downarrow$ & $\text{CHAIR}_I$ $\downarrow$ & \textit{Avg.Len} \\
\midrule
Greedy 
& 29.60$_{\pm0.89}$ & 8.03$_{\pm0.41}$ & 165.61
& 49.44$_{\pm1.57}$ & 14.19$_{\pm0.76}$ & 82.97 
& \underline{45.44}$_{\pm2.43}$ & \underline{13.07}$_{\pm0.71}$ & 92.11 \\
Beam 
& \underline{27.20}$_{\pm1.19}$ & \underline{7.20}$_{\pm0.39}$ & 174.17
& 53.60$_{\pm2.39}$ & 15.47$_{\pm0.45}$ & 87.38
& 48.68$_{\pm1.65}$ & 13.59$_{\pm0.43}$ & 95.92 \\
DoLa 
& 29.04$_{\pm1.08}$ & 7.86$_{\pm0.24}$ & 166.14
& 50.64$_{\pm2.33}$ & 14.51$_{\pm0.88}$ & 82.32 
& 46.12$_{\pm1.85}$ & 13.09$_{\pm0.90}$ & 91.80 \\
VCD 
& 31.36$_{\pm1.99}$ & 8.40$_{\pm0.79}$ & 165.43
& 51.68$_{\pm1.85}$ & 15.29$_{\pm0.83}$ & 83.03 
& 50.84$_{\pm2.41}$ & 14.51$_{\pm0.97}$ & 91.44 \\
OPERA 
& - & - & - 
& \underline{44.04}$_{\pm0.94}$ & \underline{13.23}$_{\pm0.46}$ & 75.79 
& 45.88$_{\pm2.31}$ & 13.15$_{\pm0.87}$ & 93.51 \\
CODE 
& 30.76$_{\pm0.92}$ & 8.09$_{\pm0.42}$ & 158.07
& 47.72$_{\pm0.79}$ & 14.13$_{\pm0.56}$ & 78.43 
& 50.88$_{\pm2.05}$ & 14.21$_{\pm0.92}$ & 89.62 \\
\midrule
SSL    
& \textbf{26.36}$_{\pm1.94}$ & \textbf{6.32}$_{\pm0.63}$ & 163.86
& \textbf{40.88}$_{\pm2.11}$ & \textbf{12.30}$_{\pm1.18}$ & 82.37 
& \textbf{44.04}$_{\pm3.91}$ & \textbf{12.64}$_{\pm1.38}$ & 100.26 \\
\bottomrule
\end{tabular}
}
\caption{CHAIR results on MSCOCO dataset averaged over 5 random seeds. The best and second-best results are indicated in \textbf{bold} and \underline{underlined}, respectively. \textit{Avg.Len} represents the average length of the generated
descriptions.}
\label{chair_all}
\end{table*}

We conduct experiments on three representative LVLMs: LLaVA-NeXT-8b \cite{liu2024llavanext}, LLaVA-1.5-7b \cite{llava1.5} and InstructBLIP-7b \cite{Instructblip}. These models share a modular structure comprising an image encoder, a projection module, and a language model. LLaVA-1.5 and LLaVA-NeXT use an MLP to project all image tokens into the LLM’s input space, while InstructBLIP employs a Q-Former to select a compact set of informative visual tokens, reducing redundancy. Compared to LLaVA-1.5, LLaVA-NeXT upgrades the LLM from 7b to 8b parameters and supports higher-resolution inputs for visual understanding.

\subsection{Benchmarks}

\paragraph{CHAIR.}We evaluate object hallucination using the Caption Hallucination Assessment with Image Relevance (CHAIR) metric \cite{chair}, which compares generated image captions against ground-truth annotations to detect hallucinatory objects mentioned in the captions but absent from the image. CHAIR includes two metrics at both captions level ($\text{CHAIR}_S$) and object level ($\text{CHAIR}_I$):
\begin{equation}
\begin{aligned}
\text{CHAIR}_S &= \frac{\left |\{\text{captions w/ hallucinatory objects}\}\right|}{\left |\{\text{total captions}\}\right|}, \\
\text{CHAIR}_I &= \frac{\left |\{\text{hallucinatory objects}\}\right|}{\left |\{\text{total  mentioned objects}\}\right|}.
\end{aligned}
\end{equation}
We randomly sample $500$ images from the COCO 2014 validation set \cite{MSCOCO} and conduct five runs with different random seeds. For all LVLMs, captions are generated using the prompt “Please describe this image in detail.” We report the mean and standard deviation for each metric.

\paragraph{POPE.}We further evaluate object hallucination using the POPE benchmark \cite{pope}, a question-answering dataset designed to assess the factual consistency of generated image descriptions. POPE contains $500$ images from the MSCOCO dataset \cite{MSCOCO}, each paired with binary questions of the form: “Is there a <object> in the image?” The dataset comprises three subsets—random, popular, and adversarial—which differ in their object sampling strategies. Model performance is measured using standard classification metrics: Accuracy, Precision, Recall, and F1 score. To provide an overall assessment, we report the average results across all three subsets.

\paragraph{LLaVA-Bench.}We evaluate LVLM performance using the LLaVA-Bench (In-the-Wild) benchmark \cite{llava1.5}, a comprehensive set designed to assess models across diverse and challenging visual scenarios. The benchmark includes $24$ images from varied real-world contexts, such as indoor scenes, outdoor environments, and internet memes, paired with $60$ carefully curated questions spanning open-ended QA, fine-grained descriptions, and complex reasoning. We prompt the GPT-4o model to evaluate the LVLMs’ outputs along two dimensions: factual accuracy and response detail.

\subsection{Baselines} We compare the performance of base LVLMs using greedy decoding and beam search decoding. Additionally, we also conduct a comparison between SSL and the popular training-free approaches that require neither external data nor auxiliary models. Specifically, DoLa \cite{dola} derives the next-token distribution by contrasting logits from later and earlier layers; VCD \cite{VCD} employs contrastive learning by comparing the output distributions generated from original and perturbed images; OPERA \cite{OPERA} enhances generation quality by alleviating excessive reliance on previously generated tokens during beam search; and CODE \cite{code} enhances vision-language alignment by using self-generated captions as internal references.

\subsection{Implementation Details}

We set $\gamma$ to $0.6$, $0.8$, and $0.1$ for LLaVA‑NeXT, LLaVA‑1.5, and InstructBLIP, respectively, to balance effective mitigation of hallucination with minimizing the invasiveness of state interventions. SSL is applied at $16^\text{th}$ layer for LLaVA‑NeXT, $31^\text{th}$ layer for LLaVA‑1.5 and $8^\text{th}$ layer for InstructBLIP. We faithfully replicate all baseline methods, implementing them based on their open-source codebases and configuring them according to the hyperparameters reported in the original papers. All experimental results are obtained under consistent base model, prompt, and generation parameter settings to ensure a fair comparison. For all methods involving beam search, we set the max\_new\_token to $512$ and the beam size to $5$.

\section{Results}

\textbf{CHAIR.} Table~\ref{chair_all} reports the performance of SSL on the CHAIR benchmark compared to all baseline approaches. Due to excessive memory requirements, OPERA fails to produce results on LLaVA-NeXT-8b. Notably, although the SAE from \citet{sae} were trained on LLaVA-NeXT-8b, the identified semantic directions generalize well across different model architectures. SSL consistently outperforms all baselines across all three LVLMs, while only incurring a marginal trade-off in caption length or descriptive richness.

\textbf{POPE.} As shown in Table~\ref{pope_all}, applying SSL to LVLMs with different architectures consistently improves performance on the POPE benchmark. This demonstrates the robustness of SSL in enhancing models across a spectrum of capabilities and further validates the generalizability of the semantic directions captured by our approach.

\begin{table}[t]
\centering
\small
\resizebox{\columnwidth}{!}{
\begin{tabular}{@{}c|c|c|c@{}}
\toprule
 & \multicolumn{1}{c|}{LLaVA-NeXT-8b} & \multicolumn{1}{c|}{LLaVA-1.5-7b} & \multicolumn{1}{c}{InstructBLIP-7b} \\
       & \multicolumn{1}{c|}{F1 score $\uparrow$} & \multicolumn{1}{c|}{F1 score $\uparrow$} & \multicolumn{1}{c}{F1 score $\uparrow$} \\
\midrule
Greedy  & 89.10      & 84.99 & \underline{85.37} \\
Beam    & 89.30      & 85.31     & 84.41     \\
DoLa    & \underline{89.49}     & 85.08     & 85.22     \\
VCD     & 88.91      & 84.42     & 84.68     \\
OPERA   & -      & \underline{85.46}     & 84.42     \\
CODE    & 88.93     & 84.64    & 84.81     \\
\midrule
SSL     &  \textbf{89.66}     & \textbf{85.47}     &  \textbf{85.56}    \\
\bottomrule
\end{tabular}
}
\caption{POPE results averaged over popular,
adversarial, and random splits. The best and second-best results are indicated in \textbf{bold} and \underline{underlined}, respectively.}
\label{pope_all}
\end{table}

\textbf{LLaVA-Bench.} Table~\ref{llava_bench_results} presents the evaluation results on LLaVA-Bench using GPT-4o. The results highlight the effectiveness of SSL in enhancing model accuracy across highly diverse and challenging tasks.

\begin{table}[t]
\centering
\small
\resizebox{\columnwidth}{!}{
\begin{tabular}{@{}c|c|c@{}}
\toprule
& Accuracy $\uparrow$ & Detailedness $\uparrow$ \\
\midrule
LLaVA-NeXT & 6.2891 & \textbf{6.0278} \\
LLaVA-NeXT w/ SSL & \textbf{6.3671} & 5.2667 \\
\midrule
LLaVA-1.5 & \textbf{5.3333} & \textbf{4.7000} \\
LLaVA-1.5 w/ SSL & 4.7167 & 4.5667\\
\midrule
InstructBLIP & 5.5056 & 4.1111 \\
InstructBLIP w/ SSL & \textbf{5.5823} & \textbf{4.2711}\\
\bottomrule
\end{tabular}
}
\caption{Evaluation results on the LLaVA-Bench (In-the-Wild) benchmark based on prompting GPT-4o.}
\label{llava_bench_results}
\end{table}

\subsection{Ablation Study}

\paragraph{Effectiveness of ASP.}To validate the necessity of adaptively adjusting the steering strength, we conduct an ablation study by replacing ASP with a fixed $\alpha$ steering parameter as shown in Equation~\ref{alpha} equal to the $\gamma$ value. As shown in Table~\ref{ASP}, removing the adaptive strategy ASP results in consistent performance drops across all three model architectures, highlighting the importance of ASP in effectively mitigating hallucinations.

\begin{figure}[t]
\includegraphics[width=\columnwidth]{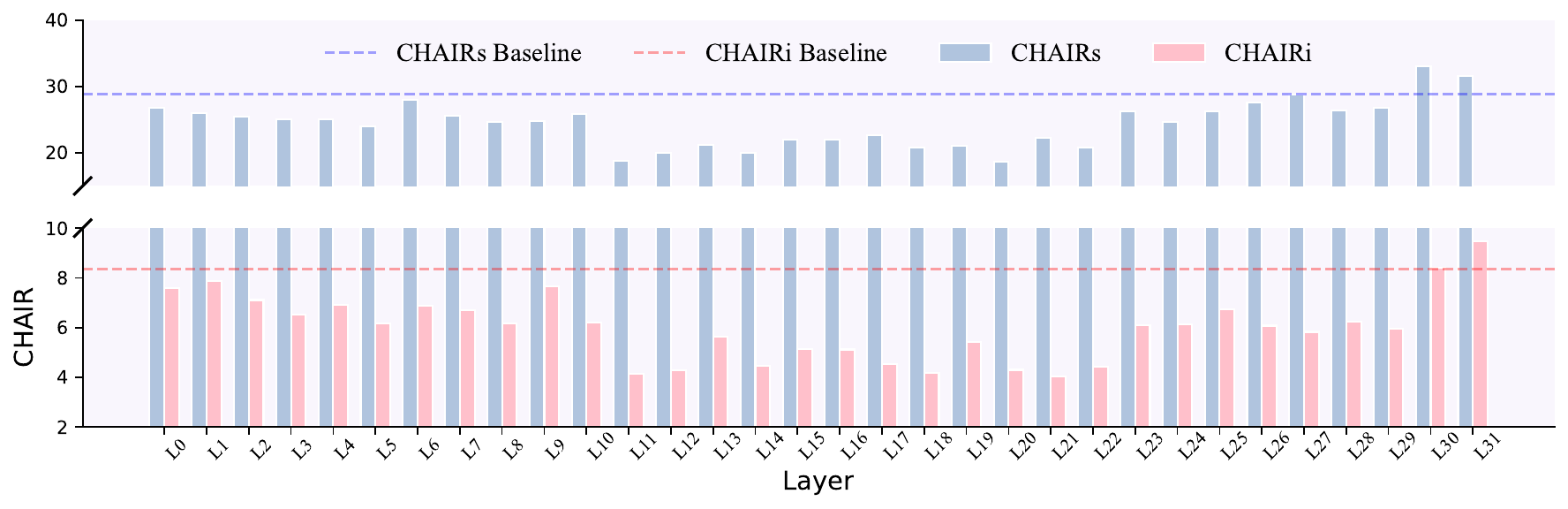}
  \caption{Results of SSL applied across different layers.}
  \label{layerablation}
\end{figure}

\paragraph{Layer Selection Ablation.}We conduct an ablation study on LLaVA-NeXT-8b to examine the impact of applying guidance at different layers. Figure~\ref{layerablation} shows that the choice of guidance layer significantly affects model performance, setting $\gamma$ to $0.8$. For LLaVA-NeXT-8b, applying SSL at middle layers yields more effective mitigation of hallucinations, with layer $15$ achieving the best performance. For further analysis on steering layer selection and scaling factor $\gamma$ settings across different models, please refer to Appendix~\ref{appendix D} and~\ref{appendix E}.

\subsection{Further Analysis}
\begin{table}[t]
\centering
\small
\resizebox{\columnwidth}{!}{
\begin{tabular}{@{}c|c|c@{}}
\toprule
& $\text{CHAIR}_S$ $\downarrow$ & $\text{CHAIR}_I$ $\downarrow$ \\
\midrule
LLaVA-NeXT w/ ASP & \textbf{26.36} & \textbf{6.32} \\
LLaVA-NeXT w/ fixed $\alpha$ & 28.40 & 7.79 \\
\midrule
LLaVA-1.5 w/ ASP & \textbf{40.88} & \textbf{12.30} \\
LLaVA-1.5 w/ fixed $\alpha$ & 47.21 & 13.12 \\
\midrule
InstructBLIP w/ ASP & \textbf{44.04} & \textbf{12.64} \\
InstructBLIP w/ fixed $\alpha$ & 45.76 & 13.12 \\
\bottomrule
\end{tabular}
}
\caption{Ablation study on the necessity of adaptively adjusting the steering parameters.}
\label{ASP}
\end{table}

\begin{figure}[t]
\subfigure[CHAIR$_S$ across models]{\includegraphics[width=0.49\columnwidth]{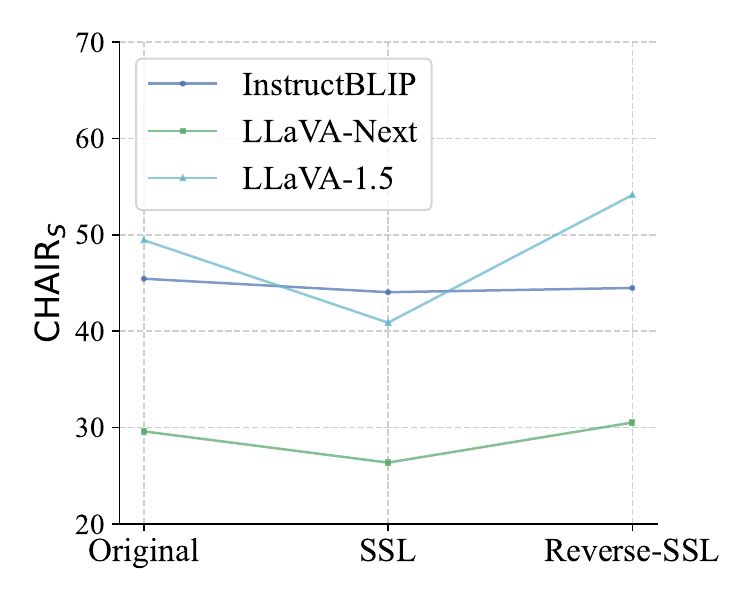}}
\subfigure[CHAIR$_I$ across models]{\includegraphics[width=0.49\columnwidth]{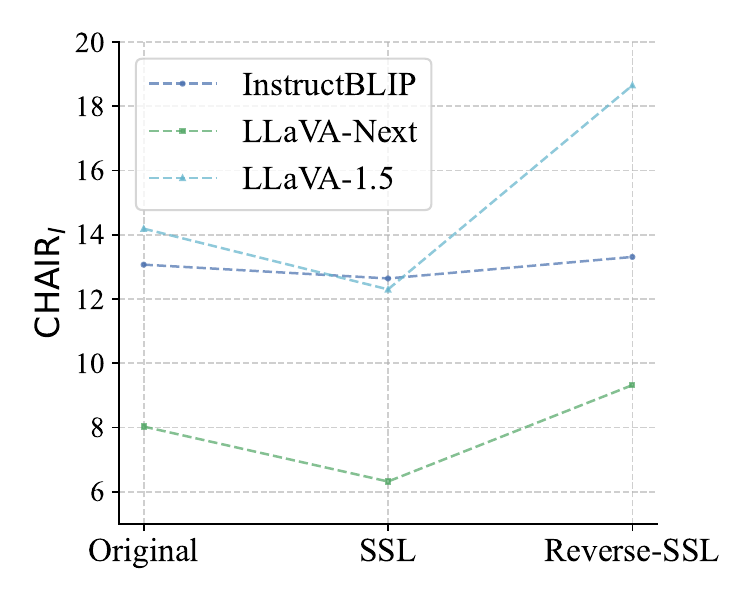}}
  \caption{CHAIR evaluation across three multimodal models—InstructBLIP-7b, LLaVA-NeXT-8b, and LLaVA-1.5-7b—under three generation settings: original, SSL, and Reverse-SSL.}
  \label{fig:hallucination_comparison}
\end{figure}

\paragraph{Analysis of Reverse-SSL for Inducing Hallucinations in LVLMs.}To further validate the effectiveness of the semantic directions we identified, we compare the CHAIR benchmark across three model architectures under three settings: original model state, SSL, and Reverse-SSL (see Appendix~\ref{appendix F} for details on Reverse-SSL). As shown in Figure~\ref{fig:hallucination_comparison}, across all three model architectures, applying Reverse-SSL significantly increases hallucinations, while standard SSL guidance consistently reduces hallucinations. To further clearly illustrate the effectiveness of SSL, we present several qualitative examples in Appendix~\ref{appendix G}.

\begin{figure}[t]
\includegraphics[width=\columnwidth]{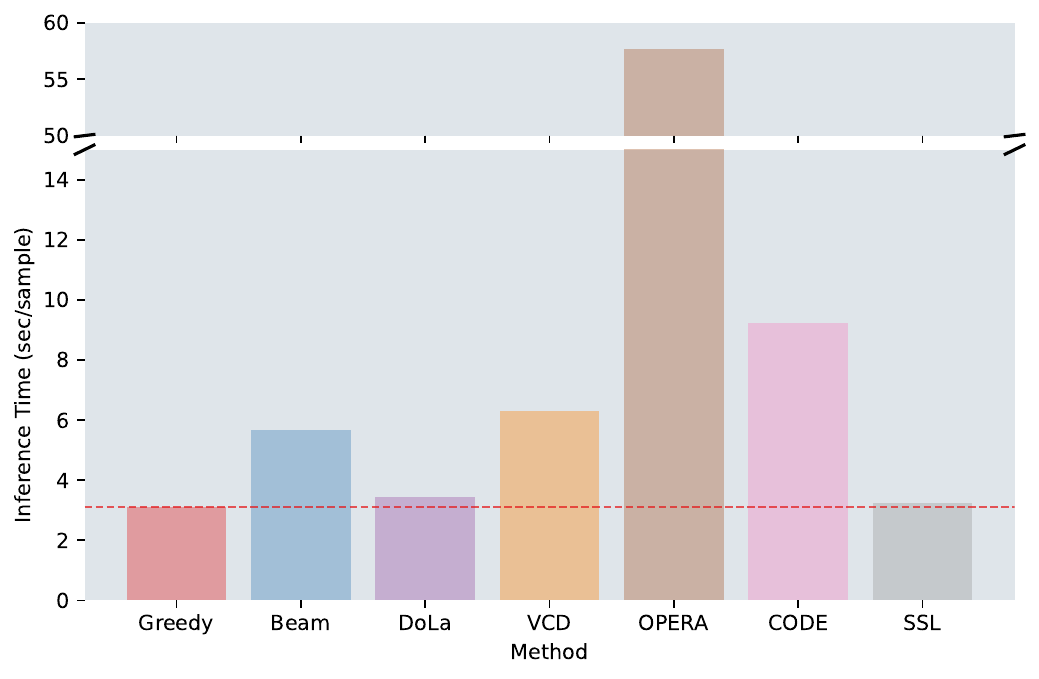}
  \caption{Comparison of inference time for different methods measured using identical hardware.}
  \label{inferencetime}
\end{figure}

\paragraph{Additional Time Analysis.}During each generation step, SSL dynamically adjusts the steering strength through a single scaling and weighting operation, introducing negligible computational overhead. Compared to the overall generation process, the additional latency introduced by SSL is minimal. A comparison of inference time between SSL and other baselines is shown in Figure~\ref{inferencetime}.

\begin{table}[t]
\centering
\small
\resizebox{\columnwidth}{!}{
\begin{tabular}{@{}c|c|c@{}}
\toprule
& $\text{CHAIR}_S$ $\downarrow$ & $\text{CHAIR}_I$ $\downarrow$ \\
\midrule
LLaVA-1.5-7b & 49.44 & 14.19 \\
LLaVA-1.5-7b w/ SSL & \textbf{40.88} & \textbf{12.30} \\
\midrule
LLaVA-Next-8b & 29.60 & 8.03 \\
LLaVA-Next-8b w/ SSL & \textbf{26.36} & \textbf{6.32} \\
\midrule
Llama-3.2-11b-Vision-Instruct & 31.40 & 7.40 \\
Llama-3.2-11b-Vision-Instruct w/ SSL & \textbf{28.92} & \textbf{6.69} \\
\bottomrule
\end{tabular}
}
\caption{CHAIR results on MSCOCO dataset averaged over 5 random seeds. The best are
indicated in \textbf{bold}.}
\label{size_chair}
\end{table}

\paragraph{Effect of Model Size on Method Performance.} To further analyze the impact of model size on our method, we conducted additional experiments on the 7B, 8B, and 11B models, evaluating SSL performance on both the CHAIR and POPE benchmarks, with results reported in Table~\ref{size_chair} and Table~\ref{size_pope}. Specifically, for the 11B model, we use Llama-3.2-11b-Vision-Instruct released by Meta Llama, where SSL is applied with $\gamma=0.4$ at the $32^{\text{th}}$ layer. The results demonstrate that SSL effectively mitigates hallucinations across models of different sizes and consistently outperforms other baseline methods.


\section{Related Works}

LVLMs refer to the phenomenon where the generated textual content is inconsistent with the visual input. This issue arises from various factors, such as dataset bias, insufficient visual perception by the encoder, and misalignment across modalities \cite{survey}. While prior studies have proposed diverse strategies to mitigate hallucination, the internal mechanisms within LVLMs that give rise to such inconsistencies remain largely underexplored.

\citet{VTI} enhance the stability of visual representations by steering latent features during generation, preventing premature degradation. \citet{jiang2025interpretingeditingvisionlanguagerepresentations} remove hallucination-related feature components through linear orthogonalization by projecting the internal image representations of vision models into the language space, thereby purifying the input and reducing hallucinations. \citet{li2025hiddenlifetokensreducing} uncover phenomena such as early activation and progressive loss of visual information in LVLMs, and propose injecting continuous visual streams during inference to compensate for these effects, significantly reducing hallucinations.

Unlike previous methods, our work directly identifies hallucinatory and faithful semantic directions using SAEs. We then dynamically adjust these directions during visual-linguistic fusion and generation to proactively reduce hallucination outputs.

Furthermore, our approach contributes to the practical interpretability of SAEs in LVLMs, demonstrating their potential for understanding and controlling internal semantic representations.

\section{Conclusion}

This work explores the relationship between the hallucination in LVLMs and their internal latent representations. We construct a residual stream dataset for hallucinatory and faithful object tokens, and use SAE to extract the semantic directions corresponding to hallucination and Factuality. Based on this insight, we propose SSL, a plug-and-play method that amplifies true semantics while suppressing potential hallucinations. Extensive experiments demonstrate that SSL outperforms existing methods. Furthermore, although the SAE was trained on LLaVA-Next, the semantic directions it extracted generalize well across different model architectures, further showcasing the potential of SAE in understanding and controlling the internal semantic representations of models.

\section*{Limitations}

Currently, the only fully open-source multi-modal SAE is provided by LLM-Labs, trained on the $25^{\text{th}}$ layer of the LLaVA-Next 8b model. As a result, our study does not include a comparison of SAEs trained on other model architectures across different multimodal models. Future work can focus on training multi-modal SAEs on various architectures to investigate whether the findings from this study generalize across different models.

\section*{Acknowledgements}

This work was supported by National Key R\&D Program of China under Grant No.2024YFC3015501, also supported by the National Natural Science Foundation of China under Grant 62576094, and Grant U24A20322.

\section*{Ethical Consideration}

In our proposed SSL method, positive steering significantly reduces hallucinations in LVLMs, while negative steering increases them. Positive steering not only improves model performance but also aligns with ethical principles such as safety and reliability. In contrast, negative guidance may lead to more hallucinations and generate false information. Therefore, such mechanisms should be applied with caution, supported by thorough validation and human oversight.
\begin{table}[t]
\centering
\small
\resizebox{\columnwidth}{!}{
\begin{tabular}{@{}c|c|c@{}}
\toprule
& $\text{Accuracy}\uparrow$ & $\text{F1 score}\uparrow$ \\
\midrule
LLaVA-1.5-7b & 83.86 & 84.99 \\
LLaVA-1.5-7b w/ SSL & \textbf{84.55} & \textbf{85.47}\\
\midrule
LLaVA-Next-8b & 89.10 & 89.10\\
LLaVA-Next-8b w/ SSL & \textbf{89.19} &\textbf{89.66} \\
\midrule
Llama-3.2-11b-Vision-Instruct & 87.26 & 87.80 \\
Llama-3.2-11b-Vision-Instruct w/ SSL & \textbf{87.48} & \textbf{88.07} \\
\bottomrule
\end{tabular}
}
\caption{POPE results averaged over popular, adversarial, and random splits. The best are indicated in \textbf{bold}.}
\label{size_pope}
\end{table}

\bibliography{custom}

\appendix

\section{Construction of Residual Stream Dataset for Hallucinatory and Faithful Object Words}
\label{appendix A}
\begin{figure*}[t]
\includegraphics[width=0.91\linewidth]{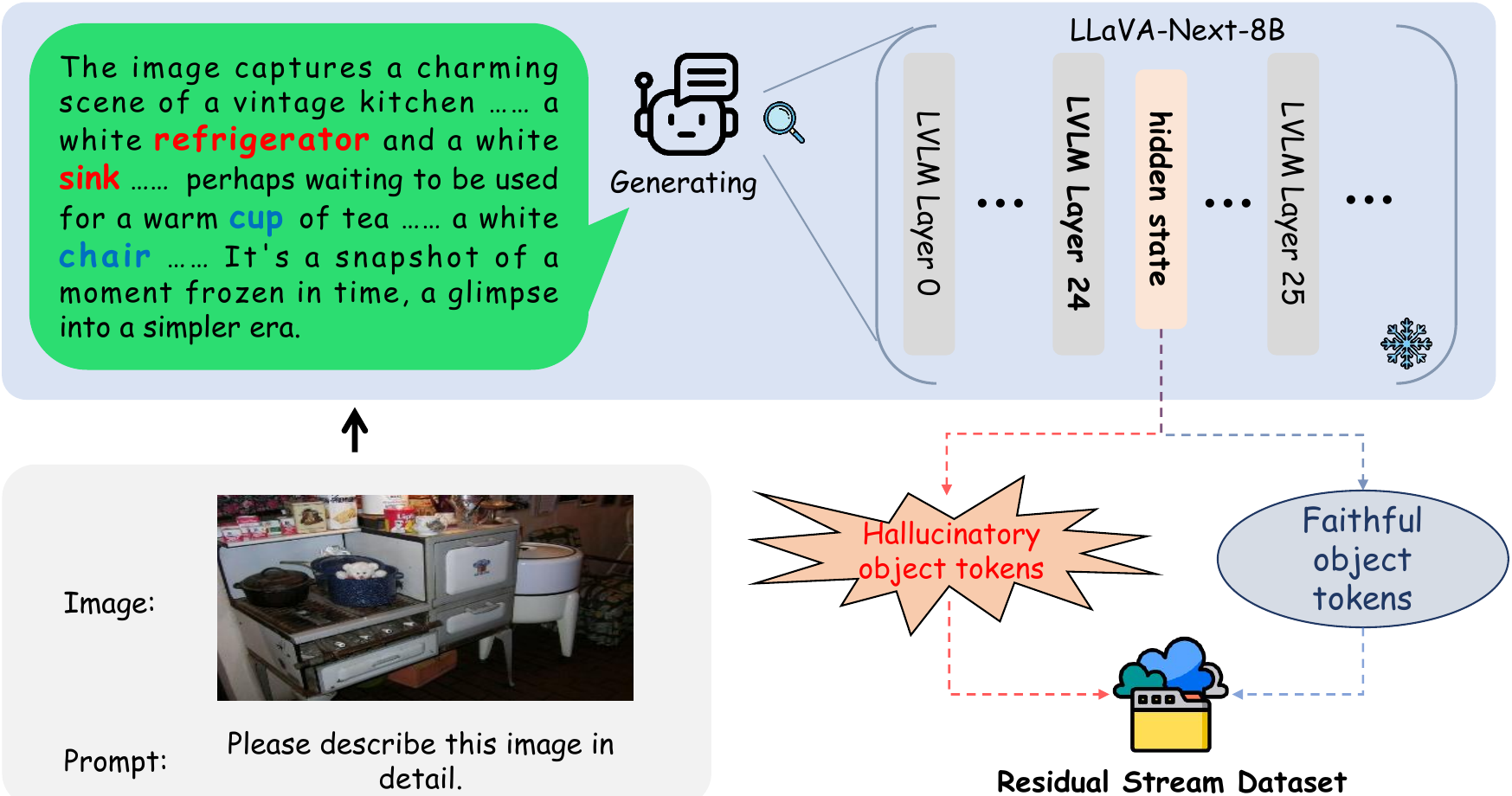}
  \caption{The process of construction of residual stream dataset for hallucinatory and faithful object words.}
  \label{dataset}
\end{figure*}

Figure~\ref{dataset} illustrates the construction process of the residual stream dataset. We begin by randomly sampling 4,000 image-text pairs from the MSCOCO dataset and extracting residual stream vectors from the 25th layer of the LLaVA-Next-8b model corresponding to object words identified as either hallucinatory or faithful during inference. It is worth noting that some object words are tokenized into multiple subword tokens by the model’s tokenizer. Given the relatively low frequency of such cases and to facilitate consistent statistical analysis, we exclude these incomplete subword instances from our dataset. For example, the word “backpack” may be split into two tokens—"back" and "pack"—by the tokenizer. Since these sub-tokens do not independently convey the complete semantic meaning of the original word, they are omitted from further analysis.

To ensure class balance, we extract an equal number of residual stream vectors for hallucinatory and faithful object words from each sample. This process results in a balanced dataset comprising 1,784 instances, with a 1:1 ratio of positive (hallucinatory) and negative (faithful) samples. We further divide the dataset into training and test sets using a 9:1 split while maintaining the class distribution in both subsets. The training set is used to identify semantic directions closely associated with hallucination, and the test set is employed to evaluate the generalizability and discriminative power of the extracted semantic features. Using sparse autoencoder analysis, we find that the direction corresponding to latent activation index 36992 is highly correlated with hallucination, whereas index 47230 aligns closely with faithful outputs.

\section{A Set of Classification Experiments Based on Logistic Regression}
\label{appendix B}

To quantitatively evaluate the discriminative power of the latent activation directions extracted by the SAE in distinguishing hallucinatory from faithful samples, we conduct a series of classification experiments based on logistic regression. Specifically, for the $i$-th sample, let $z(x_i)\in \mathbb{R}^{d_{\text{SAE}}}$ denote the SAE latent representation. From a total of $N$ samples, we extract individual latent dimensions indexed by $j$, denoted as $z_j(x_i)$, and construct five types of input features: 

Hall latent: The dimension $hall$ with the highest correlation to hallucinated object words is selected, forming a one-dimensional feature: 
\begin{equation}
X_{\text{hall}} = \left[z_{\text{hall}}(x_1),\dots,z_{\text{hall}}(x_N)\right]^T\in\mathbb{R}^{N\times1}. 
\end{equation}

Faithful latent: The dimension $faithful$ most correlated with faithful object words is selected, forming:
\begin{equation}
\begin{aligned}
&X_{\text{faithful}} \\&= \left[z_{\text{faithful}}(x_1),\dots,z_{\text{faithful}}(x_N)\right]^T\in\mathbb{R}^{N\times1}. 
\end{aligned}
\end{equation}

Random 1: A single latent dimension $r_1\sim \mathcal{U}\{0,\dots,d_{\text{SAE}}-1\}$ is randomly sampled to form a one-dimensional baseline feature $X_{r_1} = \mathbb{R}^{N\times1}$.

Random 2: Two latent dimensions $(r_1, r_2)$ are randomly sampled to construct a two-dimensional feature $X_{r_2}\in \mathbb{R}^{N\times2}$.

The corresponding label vector is $y = \left[ y^{(1)}, y^{(2)}, \dots, y^{(N)} \right]^T$, where $y^{(i)}\in\{0,1\}$, with 1 indicating a hallucinatory sample and 0 a faithful sample. All input features are standardized before being fed into the logistic regression model. The model is trained on the training set and evaluated on the held-out test set using classification accuracy and confusion matrices as evaluation metrics.

\begin{figure*}[t]
\includegraphics[width=\linewidth]{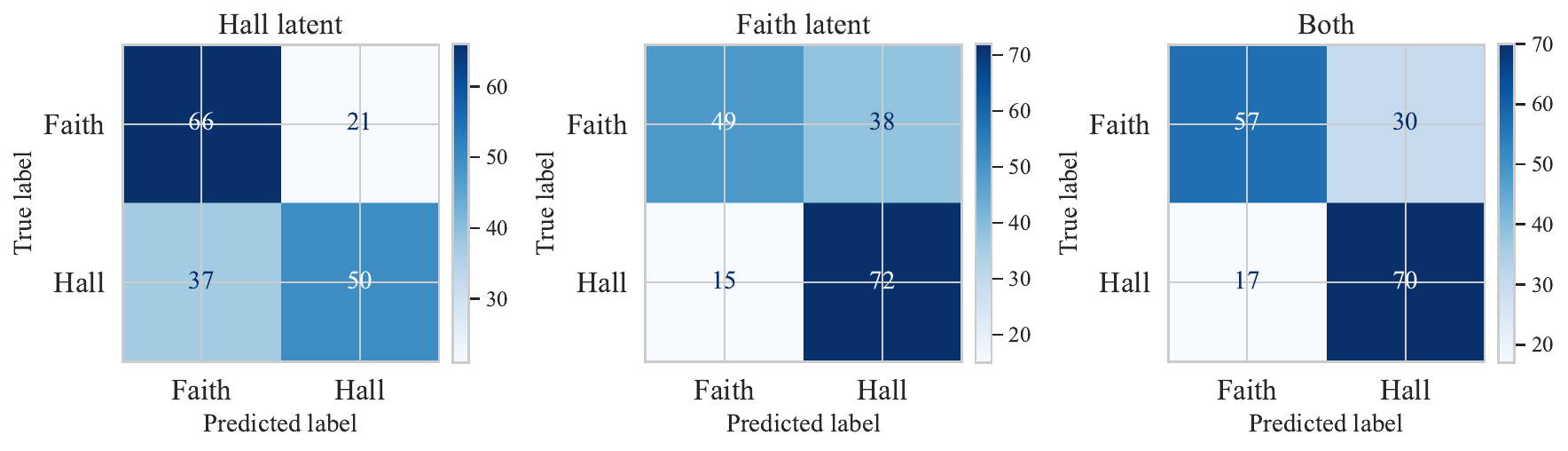}
  \caption{The confusion matrices for three main feature groups.}
  \label{confusionmatrix}
\end{figure*}
As shown in Figure~\ref{confusionmatrix}, the confusion matrices for the three main feature groups (\textit{Hall} latent, \textit{Faithful} latent, Both). The results show that the model achieves balanced performance across both positive and negative classes, with no noticeable prediction bias. Figure~\ref{logitclassification}, the performance of Random1 and Random2 approximates the random baseline ($\approx 0.5$), while using Hall latent and faithful latent individually yields classification accuracies of 66.67\% and 69.54\%, respectively. Combining the two features (Both) further improves performance to 72.99\%.

These findings suggest that the latent representations extracted by the SAE encode semantically discriminative signals for hallucination detection. Furthermore, combining hallucination and factuality-related latent directions provides complementary information that enhances classification performance.
\begin{table}[t]
\centering
\small
\resizebox{\columnwidth}{!}{
\begin{tabular}{@{}c|c|c|c@{}}
\toprule
& $\text{CHAIR}_S$ $\downarrow$ & $\text{CHAIR}_I$ $\downarrow$ & \textit{Avg.Len}\\
\midrule
LLaVA-NeXT& 29.60 & 8.03 & 165.61 \\
w/ SSL $\gamma=0.2$ & 28.36 & 7.38 & 163.20\\
w/ SSL $\gamma=0.4$ & 27.92 & 6.69 & 158.01\\
w/ SSL $\gamma=0.6$ & 26.36 & 6.32 & 163.86\\
w/ SSL $\gamma=0.8$ & 22.28 & 5.14 & 185.80\\
w/ SSL $\gamma=1.0$ & 13.28 & 4.06 & 163.62\\
w/ SSL $\gamma=1.5$ & 2.56 & 0.59 & 53.80\\
\bottomrule
\end{tabular}
}
\caption{Ablation study on the scaling factor of LLaVA-Next.}
\label{appD:llavanext}
\end{table}

\section{Boundary Between Hallucinatory and Faithful Semantic Directions}
\label{appendix C}

First, we computed the activation frequency of each latent direction across hallucinatory and faithful samples. If a direction was significantly more active in hallucinatory samples, it was labeled as hallucinatory; conversely, it was labeled as faithful. Based on this criterion, we selected the top 128 directions with the largest activation differences between the two categories (excluding dead latents) for further analysis.

We then used a Support Vector Machine (SVM) classifier in the high-dimensional space to identify a separating decision boundary between the two categories, and applied Principal Component Analysis (PCA) to project the features into two dimensions for intuitive visualization.
\begin{figure}[t]
\includegraphics[width=\columnwidth]{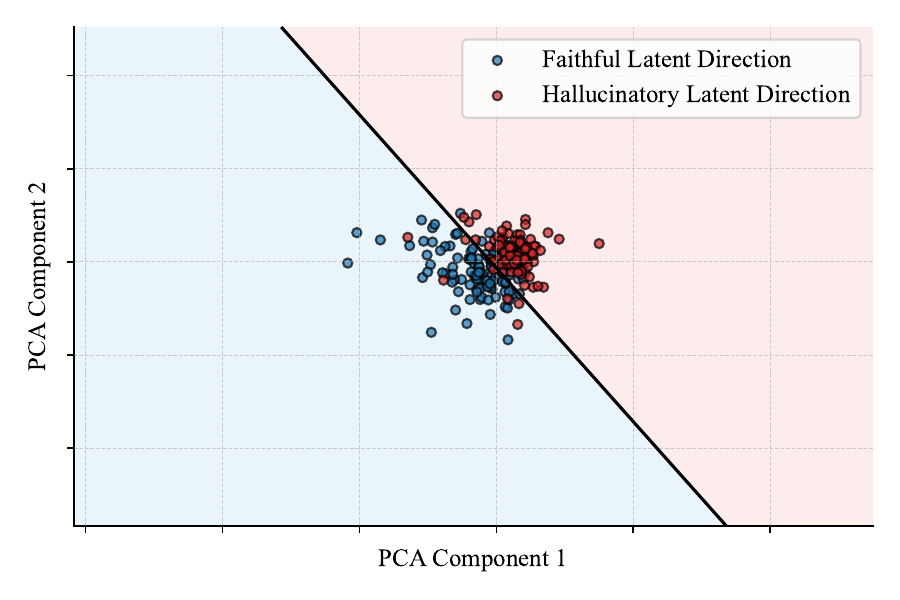}
  \caption{Visualization of the linear SVM decision boundary in the 2D PCA-reduced latent space. Blue points represent faithful latent directions, while red points represent hallucinatory latent directions. The shaded regions indicate the SVM classification regions, and the black line shows the decision boundary.}
  \label{svm}
\end{figure}
As shown in Figure~\ref {svm}, although hallucinatory and faithful semantic directions appear as relatively tight clusters in the 2D projection space, the linear decision boundary learned by the SVM still effectively separates them. This apparent overlap is mainly due to PCA prioritizing the preservation of global variance rather than discriminative features, with much of the key class-distinguishing information compressed into higher-dimensional subspaces. Nevertheless, the presence of a clear linear boundary even after dimensionality reduction indicates that hallucinatory and faithful semantic directions are inherently separable, thereby validating the existence of a semantic boundary between the two groups.

\section{Choice of Steering Layer}
\label{appendix D}
\begin{figure*}[t]
\includegraphics[width=\linewidth]{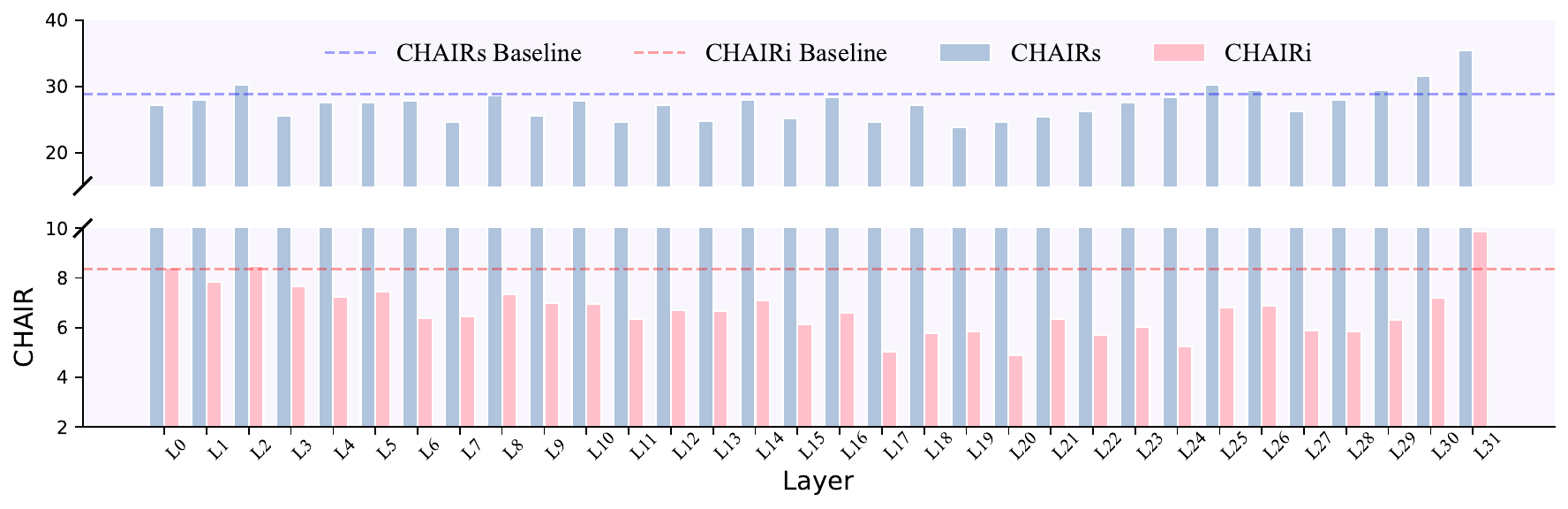}
  \caption{Ablation study on steering a specific layer of LLaVA-Next.}
  \label{appC:llavanext}
\end{figure*}
\begin{figure*}[t]
\includegraphics[width=\linewidth]{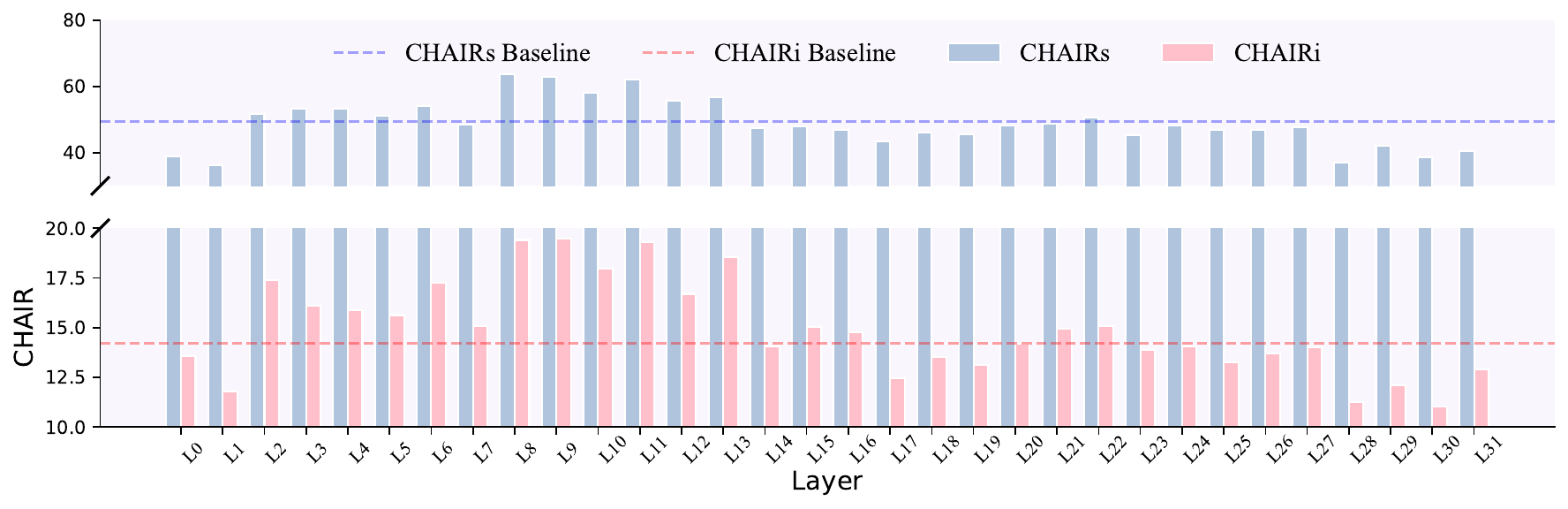}
  \caption{Ablation study on steering a specific layer of LLaVA-1.5.}
  \label{llava1.5}
\end{figure*}

\begin{figure*}[t]
\includegraphics[width=\linewidth]{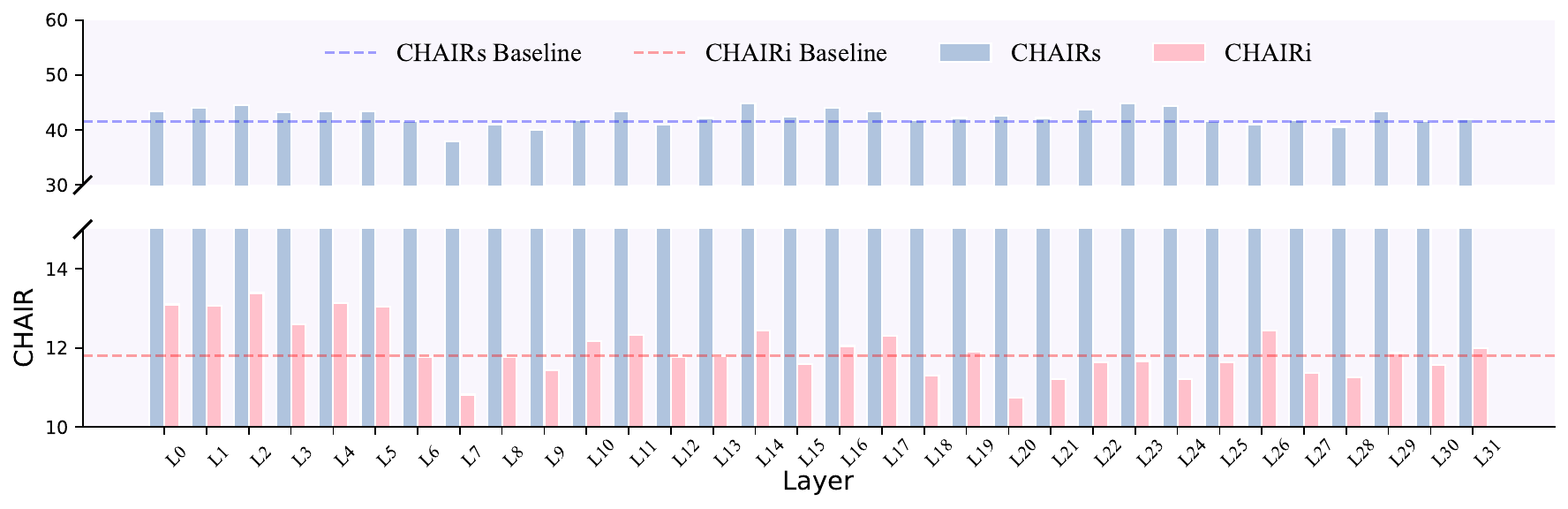}
  \caption{Ablation study on steering a specific layer of InstructBLIP.}
  \label{instructblip}
\end{figure*}
Figures~\ref{appC:llavanext},~\ref{llava1.5} and~\ref{instructblip} present the results of ablation studies investigating the effect of introducing SSL at individual layers of the LLaVA-Next ($\gamma=0.6$), LLaVA-1.5 ($\gamma=0.8)$ and InstructBLIP ($\gamma=0.1$) models, respectively. For LLaVA-Next, we observe that applying SSL at the middle layers more effectively mitigates hallucinations, consistent with the results shown in Figure~\ref{layerablation}. For LLaVA-1.5, we observe that applying SSL at either the layer1 or deeper layers consistently mitigates hallucination. This observation aligns closely with findings reported by \citet{EAH}, \citet{VHR} and \citet{chen}, who also found that layer1 or deeper layers interventions can significantly reduce hallucination in LLaVA-1.5. In contrast, for InstructBLIP, introducing SSL at shallow layers yields more substantial improvements, while deeper layer interventions contribute less noticeably to performance. We hypothesize that this is attributable to architectural and training differences in InstructBLIP, specifically, its shallow layers may already perform substantial cross-modal alignment early in the pipeline, making early-stage semantic guidance more impactful on overall generation quality. A deeper analysis of the layer-specific mechanisms in different multimodal architectures is left for future work.

\begin{algorithm}[t]
\caption{Reverse-SSL}
\label{alg:rssl}
\KwIn{Scaling factor $\gamma$; steering layer $l_s$; semantic directions $d_{\text{hall}}$, $d_{\text{faithful}}$; residual stream at layer $l_s$: $[X_{l_s,s}, X_{l_s,t}, X_{l_s,v}, X_{l_s,o}^{<t}]$}
\If{$L = l_s$}{
    \For{token $x$ in residual stream}{
        \uIf{$x \in X_{l_s,v}$}{
            $x \leftarrow x - \gamma \cdot \frac{\|x\|}{\|d_{\text{faithful}}\| + \epsilon} \cdot d_{\text{faithful}}$
        }
        \uElseIf{$x \in X_{l_s,o}^{<t}$}{
            $x \leftarrow x + \gamma \cdot \frac{\|x\|}{\|d_{\text{hall}}\| + \epsilon} \cdot d_{\text{hall}}$
        }
        \Else{
            \tcp{System and prompt tokens remain unchanged}
        }
    }
}
\end{algorithm}

\section{Choice of Scaling Factor}
\label{appendix E}

Tables~\ref{appD:llavanext},~\ref{appD:llava1.5}, and \ref{appD:InstructBLIP} report the ablation results on the effect of the scaling factor $\gamma$ in the SSL. For the LLaVA series of models, setting $\gamma$ to 0.6 or 0.8 effectively reduces hallucinations, indicating that moderate levels of semantic intervention are beneficial. However, when $\gamma > 1.0$, the supervision becomes overly aggressive, disrupting the model’s behavior and leading to abnormal hallucination metrics. In contrast, for InstructBLIP, larger values similarly result in performance degradation. These findings highlight the importance of carefully calibrating the intensity of semantic guidance to balance model control and generation quality across different architectures.
\begin{table}[t]
\centering
\small
\resizebox{\columnwidth}{!}{
\begin{tabular}{@{}c|c|c|c@{}}
\toprule
& $\text{CHAIR}_S$ $\downarrow$ & $\text{CHAIR}_I$ $\downarrow$ & \textit{Avg.Len}\\
\midrule
LLaVA-1.5& 49.44 & 14.19 & 82.97 \\
w/ SSL $\gamma=0.2$ & 48.16 & 14.10 & 83.45\\
w/ SSL $\gamma=0.4$ & 47.16 & 14.04 & 83.24\\
w/ SSL $\gamma=0.6$ & 45.96 & 13.36 & 83.63\\
w/ SSL $\gamma=0.8$ & 40.88 & 12.30 & 82.37\\
w/ SSL $\gamma=1.0$ & 33.80 & 10.18 & 86.02\\
w/ SSL $\gamma=1.5$ & 17.16 & 7.66 & 275.16\\
\bottomrule
\end{tabular}
}
\caption{Ablation study on the scaling factor of LLaVA-1.5.}
\label{appD:llava1.5}
\end{table}
\begin{table}[t]
\centering
\small
\resizebox{\columnwidth}{!}{
\begin{tabular}{@{}c|c|c|c@{}}
\toprule
& $\text{CHAIR}_S$ $\downarrow$ & $\text{CHAIR}_I$ $\downarrow$ & \textit{Avg.Len}\\
\midrule
InstructBLIP& 45.44 & 13.07 & 92.11 \\
w/ SSL $\gamma=0.1$ & 44.04 & 12.64 & 102.86\\
w/ SSL $\gamma=0.2$ & 41.92 & 12.10 & 100.26\\
w/ SSL $\gamma=0.3$ & 38.40 & 11.67 & 105.38\\
w/ SSL $\gamma=0.4$ & 37.21 & 11.45 & 119.39\\
w/ SSL $\gamma=0.5$ & 8.61 & 10.55 & 44.49\\
w/ SSL $\gamma=0.6$ & 3.42 & 14.02 & 52.14\\
\bottomrule
\end{tabular}
}
\caption{Ablation study on the scaling factor of InstructBLIP.}
\label{appD:InstructBLIP}
\end{table}

\section{Reverse-SSL Approach}
\label{appendix F}

To further validate the efficacy of the identified semantic directions, we extend the standard SSL by introducing Reverse Steering LVLMs via SAEs Latents (Reverse-SSL), an approach that deliberately induces the model to generate more hallucinations. Specifically, at each residual flow layer, we divide the input sequence into four contiguous segments: system tokens, prompt tokens, visual tokens, and output tokens. During the visual-token stage, we inject a specific reverse-direction vector that deliberately shifts the visual features away from the true image semantics. At the onset of autoregressive language generation, we inject the specific reverse-direction to amplify the previously distorted visual signal, thereby biasing subsequent text outputs toward content that is either factually incorrect or substantially divergent from the original prompt. The complete algorithmic procedure for Reverse-SSL is presented in Algorithm~\ref {alg:rssl}.


\section{More Qualitative Results}
\label{appendix G}

\begin{figure*}[t]
\includegraphics[width=\linewidth]{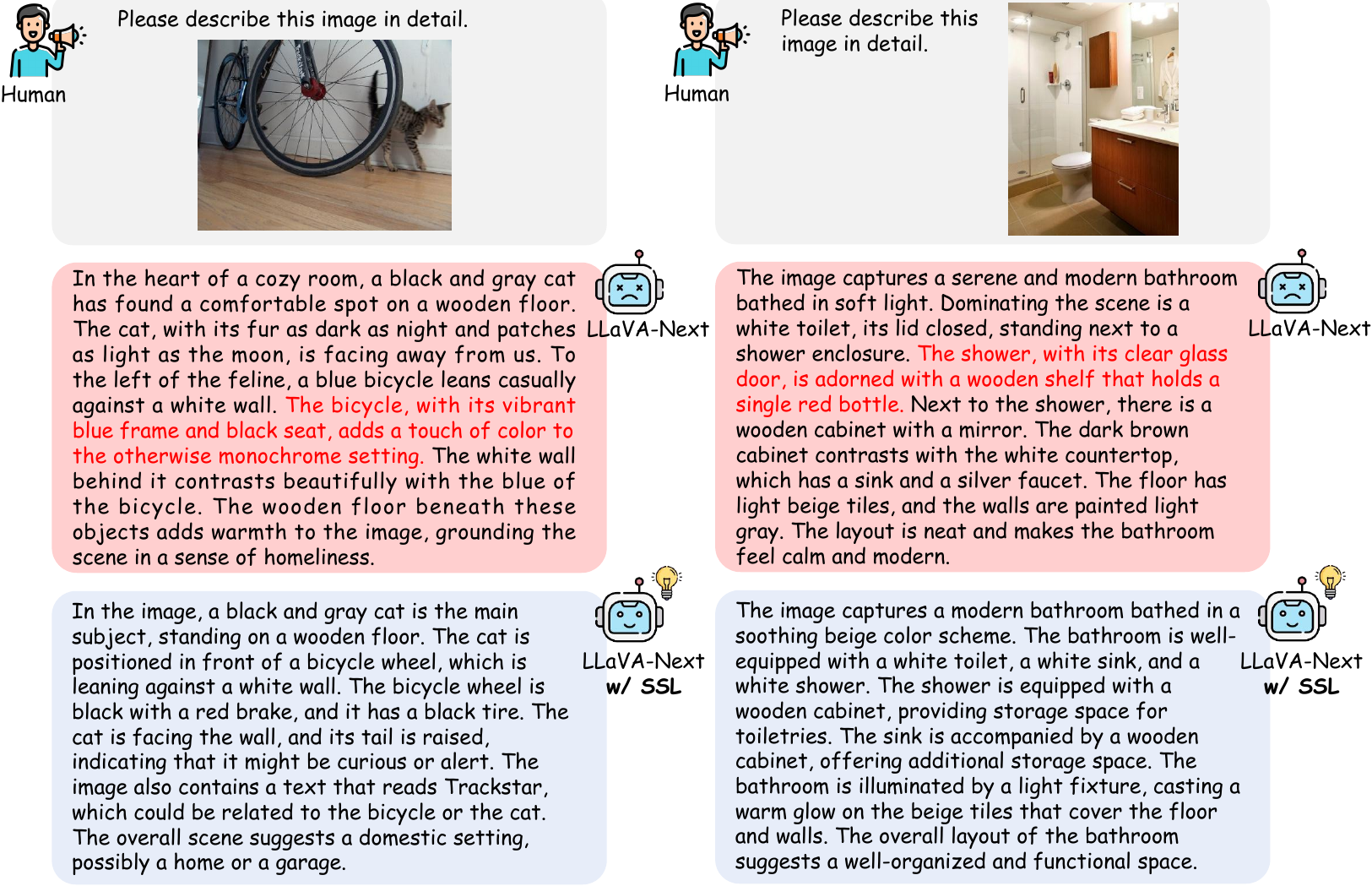}
  \caption{More examples of SSL in reducing hallucinated content in LLaVA-Next.}
  \label{appF:llavanext}
\end{figure*}

\begin{figure*}[t]
\includegraphics[width=\linewidth]{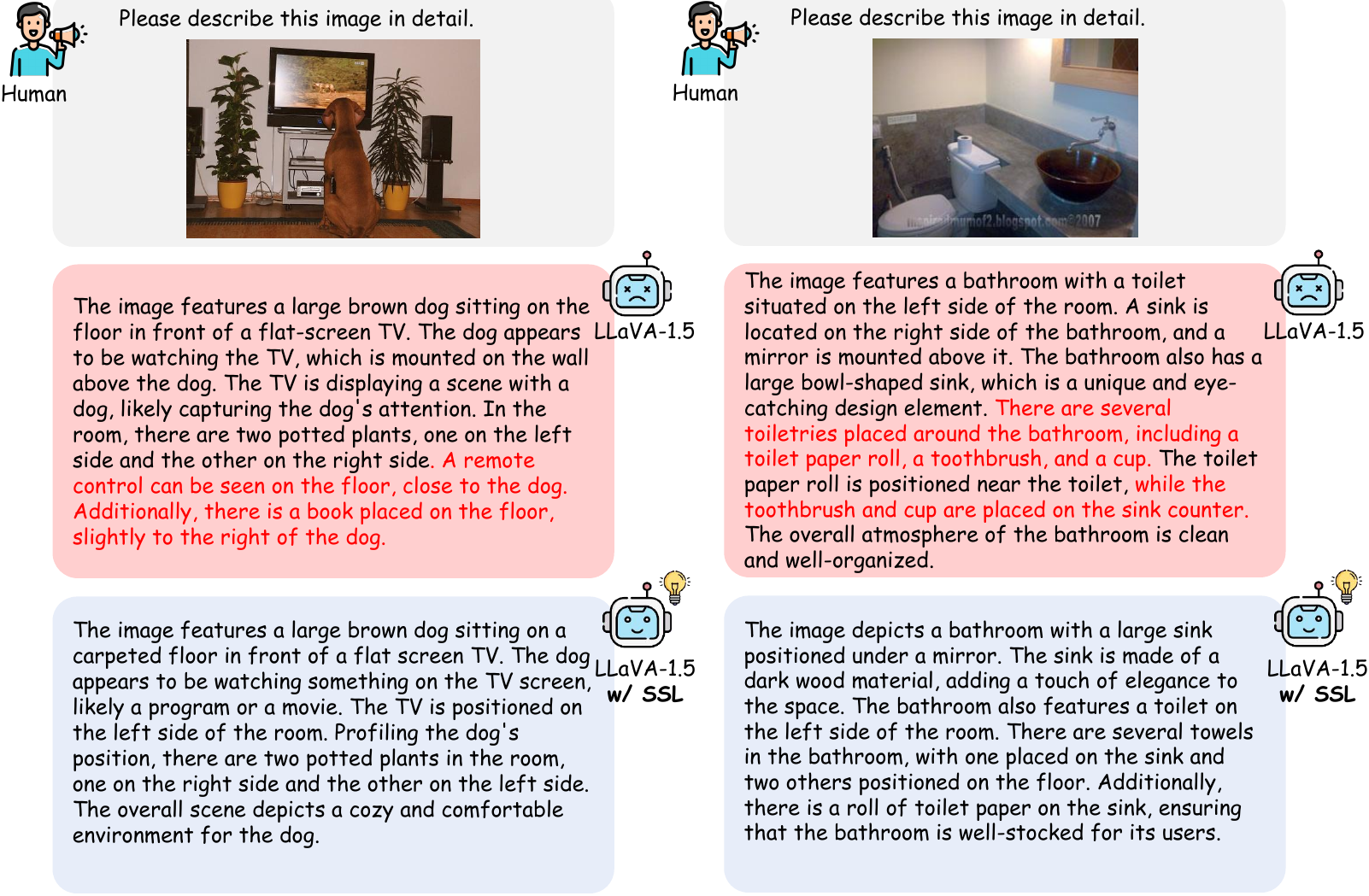}
  \caption{More examples of SSL in reducing hallucinated content in LLaVA-1.5.}
  \label{appF:llava1.5}
\end{figure*}

\begin{figure*}[t]
\includegraphics[width=\linewidth]{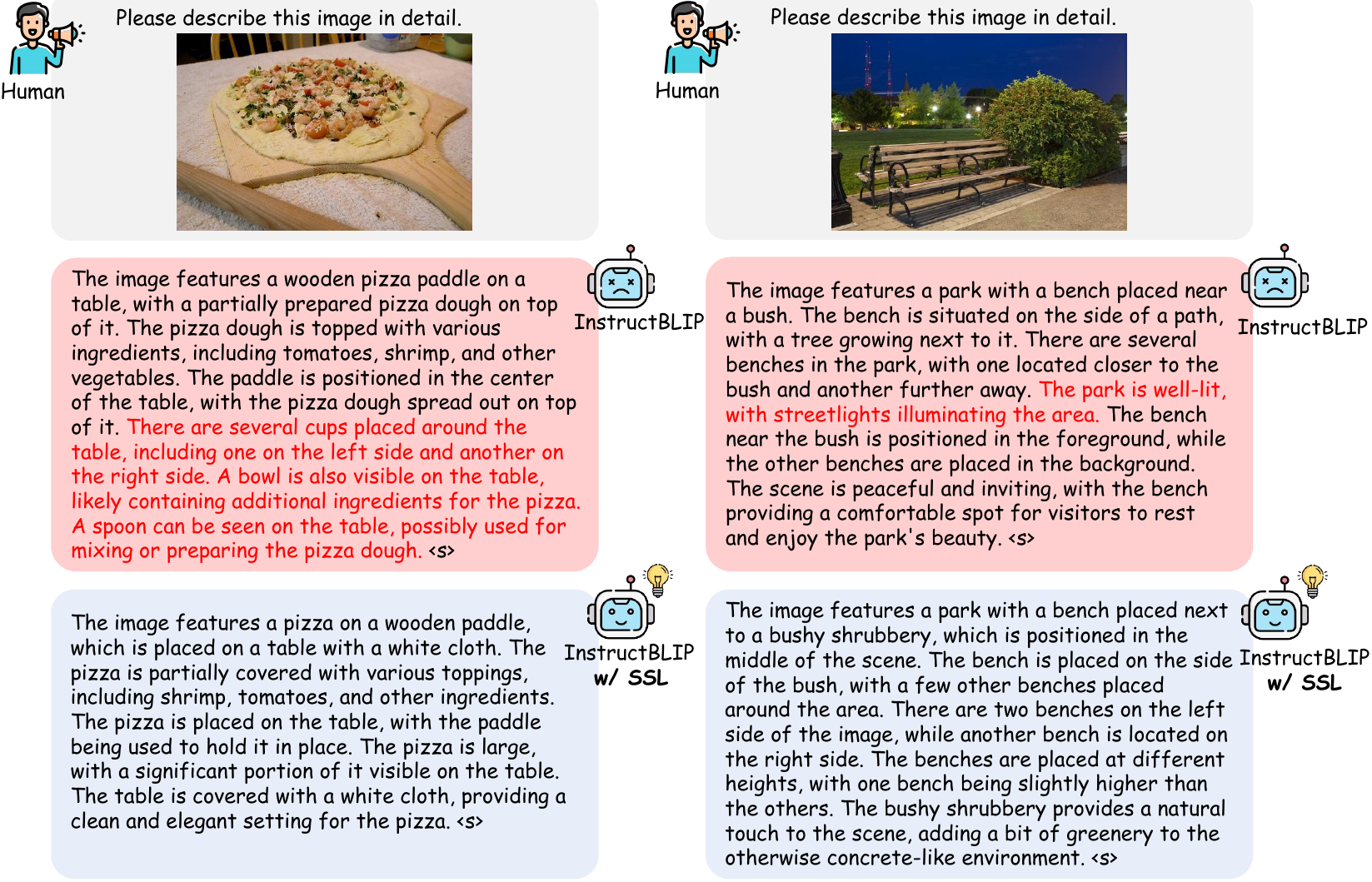}
  \caption{More examples of SSL in reducing hallucinated content in InstructBLIP.}
  \label{appF:instructblip}
\end{figure*}

Figures~\ref{appF:llavanext},~\ref{appF:llava1.5}, and~\ref{appF:instructblip} present additional qualitative examples on the LLaVA-Next, LLaVA-1.5, and InstructBLIP models, respectively, to demonstrate the effectiveness of our proposed SSL approach in mitigating hallucinated objects. With the integration of SSL, the generated descriptions by LVLMs exhibit improved fidelity to the visual content while maintaining the richness and informativeness of the language output.

\section{Details on the GPT-4o Evaluation}
\label{appendix H}
\begin{figure*}[t]
\includegraphics[width=0.91\linewidth]{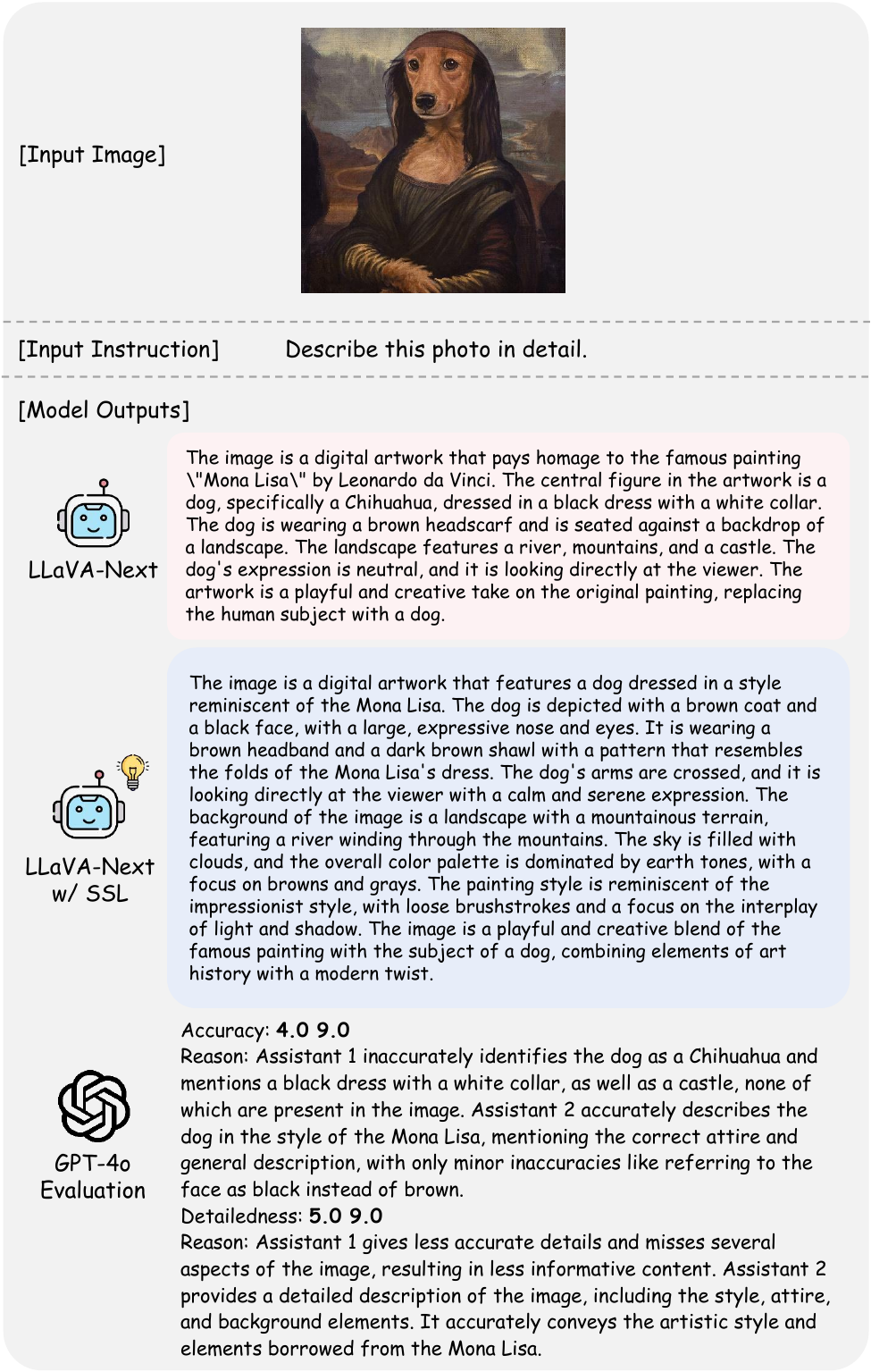}
  \caption{The performance of SSL on reducing hallucinations on LLaVA-Next-8b.}
  \label{appH:llavanext}
\end{figure*}

\begin{figure*}[t]
\includegraphics[width=0.91\linewidth]{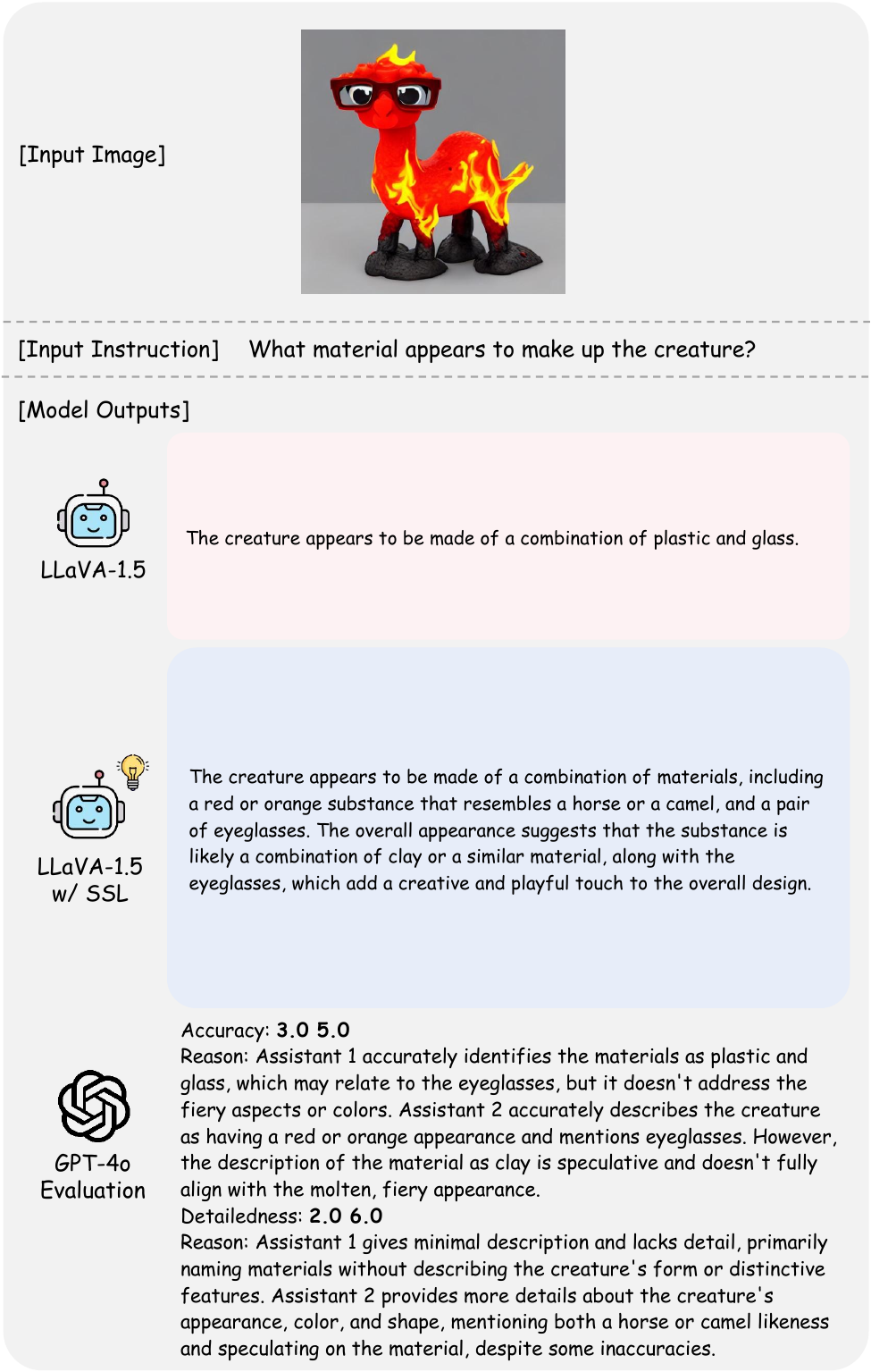}
  \caption{The performance of SSL on reducing hallucinations on LLaVA-1.5-7b.}
  \label{appH:llava1.5}
\end{figure*}

\begin{figure*}[t]
\includegraphics[width=0.91\linewidth]{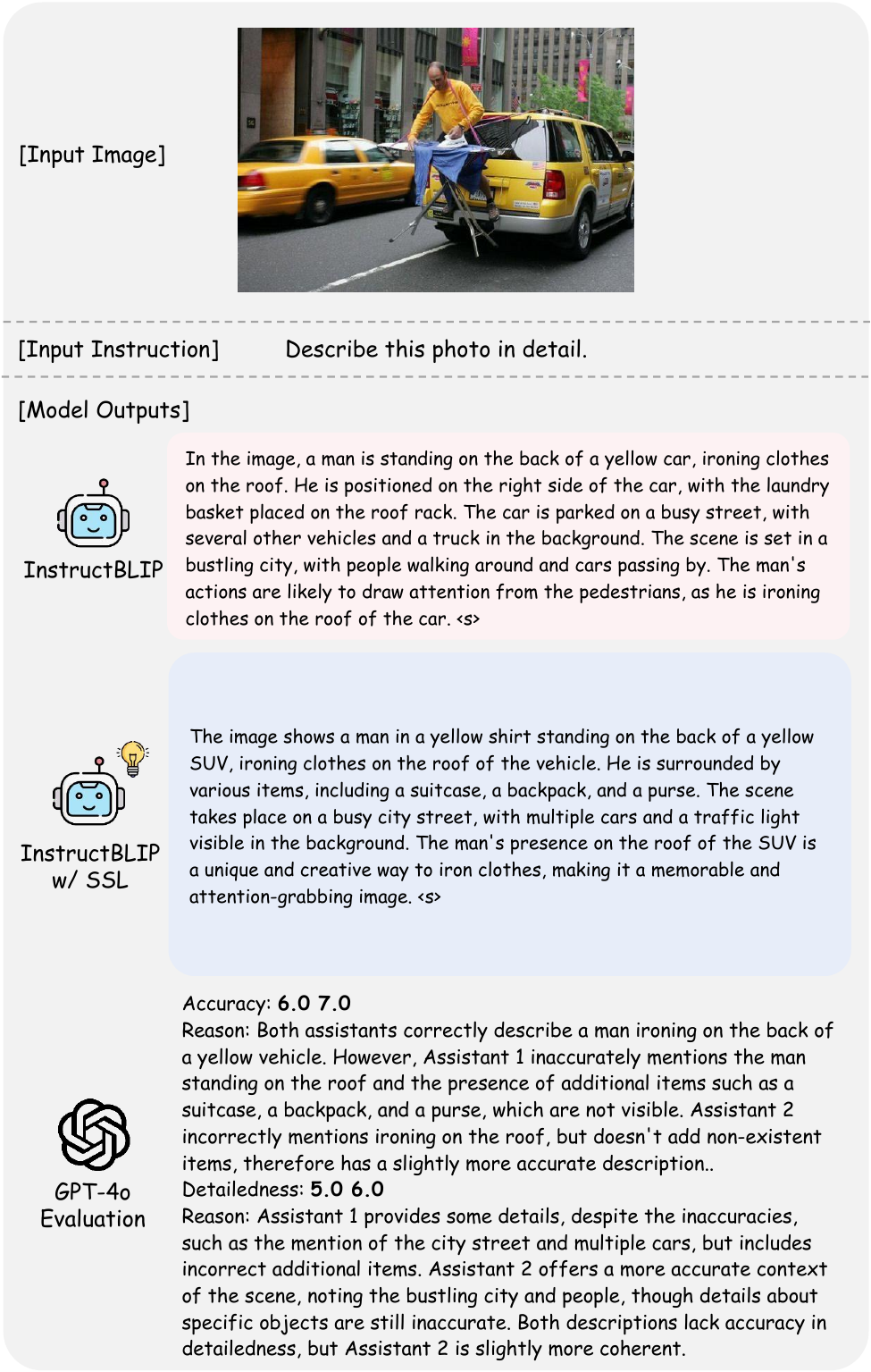}
  \caption{The performance of SSL on reducing hallucinations on InstructBLIP-7b.}
  \label{appH:instructblip}
\end{figure*}

To evaluate the performance of LVLMs on the LLaVA-Bench benchmark, we adopt GPT-4o as the reference evaluator. Following the template provided in Table~\ref{GPT-4o} of \citet{gong2024damrodiveattentionmechanism}, each evaluation instance includes the original image, the base output of the LVLM, and its SSL-enhanced counterpart. The evaluation focuses on both the accuracy and fineness of the generated responses. To mitigate potential biases caused by output order, we randomly swap the positions of the two outputs with a probability of 0.5 before each evaluation. Each sample is evaluated four times to compute an average score. Figures~\ref{appH:llavanext}, ~\ref{appH:llava1.5}, and ~\ref{appH:instructblip} illustrate representative evaluation examples based on three model architectures: LLaVA-Next, LLaVA-1.5, and InstructBLIP, respectively.

\section{License}
The open-source data utilized in this work was employed exclusively for academic research, consistent with the original intended usage. All the used intellectual artifacts’ license allows for academic usage.

\begin{table*}[t]
\centering
\begin{tabular}{p{0.95\linewidth}}
\toprule
\textbf{GPT-4o Prompt} \\
\midrule
You are required to score the performance of two AI assistants in describing a given image. You should pay extra attention to the hallucination, which refers to the part of descriptions that are inconsistent with the image content, such as claiming the existence of something not present in the image or describing incorrectly in terms of the counts, positions, or colors of objects in the image. Please rate the responses of the assistants on a scale of 1 to 10, where a higher score indicates better performance, according to the following criteria: \\
\textbf{1. Accuracy}: whether the response is accurate with respect to the image content. Responses with fewer hallucinations should be given higher scores.\\
\textbf{2. Detailedness}: whether the response is rich in necessary details. Note that hallucinated descriptions should not count as necessary details.\\[0.5em]
Please output the scores for each criterion, containing only two values indicating the scores for Assistant 1 and 2, respectively. The two scores are separated by a space. Following the scores, please provide an explanation of your evaluation, avoiding any potential bias and ensuring that the order in which the responses were presented does not affect your judgment.

[Assistant 1] \\

\{\} 

[End of Assistant 1] 

\\

[Assistant 2] 

\{\} 

[End of Assistant 2] 

\\
Output format: \\
Accuracy: \\
Reason: \\
Detailedness: \\
Reason: \\
\bottomrule
\end{tabular}
\caption{The prompt used for GPT-4o evaluation.}
\label{GPT-4o}
\end{table*}

\end{document}